\documentclass[11pt]{article}
\usepackage[preprint]{acl}
\usepackage{times}
\usepackage{latexsym}
\usepackage[T1]{fontenc}
\usepackage[utf8]{inputenc}
\usepackage{microtype}
\usepackage{amsmath,amssymb,amsthm,graphicx,booktabs}
\usepackage{colortbl}
\usepackage[most]{tcolorbox}
\DeclareCaptionType{anchoralgorithm}[Algorithm][List of Algorithms]
\usepackage{pifont}
\usepackage{enumitem}
\usepackage{needspace}
\usepackage{xspace}
\newtheorem{proposition}{Proposition}
\newtheorem{assumption}{Assumption}

\newcommand{\methodname}{%
  \texorpdfstring{{\fontfamily{lmtt}\selectfont\textbf{Anchor}}}{Anchor}\xspace}
\newcommand{\localizer}{\mathcal{L}_{\mathrm{loc}}}

\newcommand{\bankmaster}{\mathcal{B}_{\mathrm{master}}}
\usepackage{fontawesome5}
\IfFileExists{fontawesome5.map}{\pdfmapfile{+fontawesome5.map}}{}
\definecolor{rq1}{HTML}{285B87}
\definecolor{rq2}{HTML}{72538B}
\definecolor{rq3}{HTML}{197568}
\definecolor{rq4}{HTML}{966122}
\definecolor{anchorInk}{HTML}{263C50}
\definecolor{oursgray}{HTML}{E5F2EF}
\definecolor{groupgray}{HTML}{F5F7F9}
\renewcommand{\topfraction}{.9}

\renewcommand{\textfraction}{.08}
\DeclareRobustCommand{\papericon}[2]{%
  \texorpdfstring{\textcolor{#1}{\scalebox{.84}{\faIcon{#2}}}}{}}
\newcommand{\localizermark}{\papericon{rq1}{crosshairs}}
\newcommand{\amplitudermark}{\papericon{rq4}{wave-square}}
\newcommand{\aspectermark}{\papericon{rq2}{compass}}
\newcommand{\bankmastermark}{\papericon{rq3}{database}}

\newcommand{\anchormark}{\papericon{rq3}{anchor}}
\newcommand{\alfmark}{\papericon{rq1}{home}}
\newcommand{\webmark}{\papericon{rq1}{shopping-cart}}
\newcommand{\successmark}{\papericon{rq3}{check-circle}}
\newcommand{\failuremark}{\papericon{rq4}{undo-alt}}
\DeclareRobustCommand{\rqmark}[1]{%
  \texorpdfstring{\mbox{\textbf{RQ#1}}}{RQ#1}}
\newlist{rqlist}{itemize}{1}
\setlist[rqlist]{label={},leftmargin=0pt,topsep=2pt,itemsep=1.5pt,
  parsep=0pt,partopsep=0pt}

\newcommand{\resultobs}[2]{%
  \par\addvspace{4pt}\noindent
  \mbox{\textbf{Answer #1.}}\enspace\textbf{#2}\ }
\newcommand{\best}[1]{\textbf{#1}}
\newcommand{\second}[1]{\underline{#1}}
\newcommand{\runinhead}[1]{\paragraph{\normalfont\bfseries #1}}
\newcommand{\introhead}[1]{%
  \par\addvspace{3pt}\noindent\textbf{#1}\enspace\ignorespaces}
\newcommand{\tablegroup}[2]{\papericon{anchorInk}{#1}\hspace{.4em}\textit{#2}}
\newcommand{\modelgroup}[1]{\raisebox{-.18em}{\includegraphics[height=1.05em]{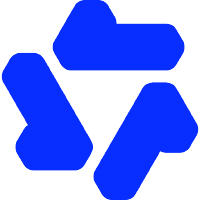}}\hspace{.45em}\textbf{#1}}
\newcommand{\tableheader}[1]{\textbf{#1}}
\newcommand{\tablemark}[1]{\makebox[1.1em][c]{#1}\hspace{.3em}}
\newcommand{\anchorrowname}{\tablemark{\anchormark}\methodname{} (ours)}
\newcommand{\anchortablestyle}{%
  \arrayrulecolor{black}%
  \setlength{\heavyrulewidth}{.8pt}%
  \setlength{\lightrulewidth}{.35pt}%
  \setlength{\cmidrulewidth}{.3pt}%
  \setlength{\aboverulesep}{1.8pt}%
  \setlength{\belowrulesep}{1.8pt}}
\newtcolorbox{anchorquestion}{enhanced,boxrule=0pt,frame hidden,
  colback=rq1!4,borderline west={1.4pt}{0pt}{rq1},sharp corners,
  left=5pt,right=4pt,top=3pt,bottom=3pt,boxsep=0pt,
  before skip=5pt,after skip=5pt}

\newcommand{\ablnobehavior}{\methodname{} w/o Amplituder \& Aspecter}
\newcommand{\ablnoaspecter}{\methodname{} w/o Aspecter}
\newcommand{\ablnofailure}{\methodname{} w/o Failure Corrections}
\newcommand{\ablnosuccess}{\methodname{} w/o Success Maintain}
\title{From Reflection to Co-Evolution:\\
Self-Evolving Agents with Locally Verified Memory}
\author{%
  \normalfont\normalsize Xinxiang Yin\textsuperscript{1}, Xinglang Zhang\textsuperscript{2}, Haodong Chen\textsuperscript{3}, Zhiqing Cui\textsuperscript{3}, Weidong Tang\textsuperscript{4},\\
  \normalfont\normalsize Yuxuan Wang\textsuperscript{5}, Junrui Wang\textsuperscript{6}, Jingyi Zhang\textsuperscript{1}, Jianfeng Zhang\textsuperscript{7}, Jian Qian\textsuperscript{1}\\[5pt]
  \normalfont\small\textsuperscript{1}The University of Hong Kong\quad
  \textsuperscript{2}Huazhong University of Science and Technology\\
  \normalfont\small\textsuperscript{3}The Hong Kong University of Science and Technology\\
  \normalfont\small\textsuperscript{4}National University of Singapore\quad
  \textsuperscript{5}Peking University\\
  \normalfont\small\textsuperscript{6}Xi'an Jiaotong University\quad
  \textsuperscript{7}Huawei Noah's Ark Lab\\[4pt]
  \normalfont\small\texttt{\{xinxiangyin24\}@gmail.com}\quad\texttt{\{jianqian\}@hku.hk}}
\hypersetup{pdfauthor={Xinxiang Yin, Xinglang Zhang, Haodong Chen, Zhiqing Cui, Weidong Tang, Yuxuan Wang, Junrui Wang, Jingyi Zhang, Jianfeng Zhang, Jian Qian},
  pdftitle={From Reflection to Co-Evolution: Self-Evolving Agents with Locally Verified Memory}}

\begin{document}
\maketitle
\raggedbottom

\begin{abstract}
LLM agents have made impressive progress in coding and mathematical reasoning, motivating growing interest in learning from long-horizon interaction. Memory supports this process, but plausible lessons can be redundant or reinforce inappropriate actions. Can a memory's source decision provide a practical test for whether to retain it? We localize consequential trajectory steps, extract candidate memories, and replay each source decision with and without its candidate under the same context. This reveals two properties: most candidates barely shift the probed distribution, and both successful and failed source actions can be reinforced or suppressed. These observations motivate \methodname{}, which tests shift magnitude and source-role alignment before retention. Admitted memories guide later trajectories, enabling experience-driven improvement under a frozen policy. On held-out ALFWorld and WebShop tasks, \methodname{} improves success rate and mean score by up to 19.9\% and 17.2\%, respectively, relative to the strongest baseline for each backbone. With Qwen3.5-27B, it retains 41\% fewer ALFWorld cards than unfiltered retention and outperforms all four WebShop cost-comparison baselines while using fewer tokens.
\end{abstract}

\section{Introduction}
\label{sec:intro}

\begin{figure*}[!t]
    \centering
    \includegraphics[width=.94\textwidth]{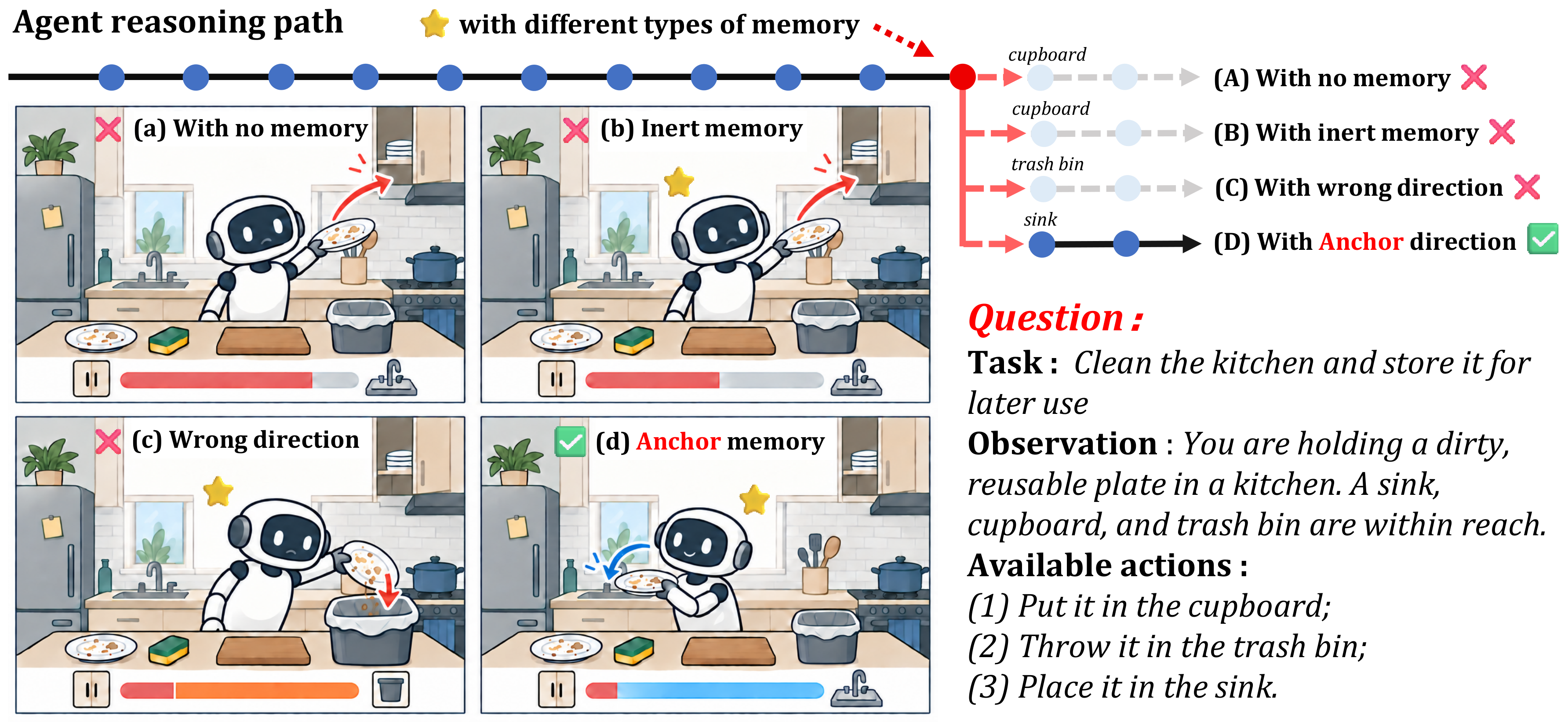}
    \caption{\textbf{Memory can change the next action.} In this schematic kitchen example, putting a dirty reusable plate in the sink is preferable to the cupboard or trash bin. \methodname{} screens for a role-aligned change.}
    \label{fig:anchor-teaser}
\end{figure*}

Large language model (LLM) applications span coding~\citep{shi2026codehackerautomatedtestcase,yang2024sweagentagentcomputerinterfacesenable}, mathematical and logical reasoning~\citep{ren2025deepseekproverv2advancingformalmathematical,zhang2026logicalphasetransitionsunderstanding,zhang2026semioticlogicalhexagontheory}, scientific investigation~\citep{zhang2026valgagenticmltheory,cui2026earthversebenchmarkingscientificagents}, multimodal reasoning and generation~\citep{chen2025finequestadaptiveknowledgeassistedsports,chen2026showdonttellmorphing}, social simulation~\citep{zhang2026gas3comprehensivesocialnetwork,zhang2026intervensiminterventionawaresocialnetwork,zhang2026couplingmacrodynamicsmicro,tang2026grouptom}, and cognitive training~\citep{shi2026letgamesllmpoweredgamifiedapproach}. Beyond single tasks, recent work studies how agents improve via long-horizon interaction and experience reuse~\citep{wang2023voyageropenendedembodiedagent,shinn2023reflexionlanguageagentsverbal,tang2026trinitycapabilitydrivenrecursiveselfimprovement}. Memory provides a direct mechanism: Reflexion, ExpeL, and ReasoningBank extract reusable lessons from trajectories without updating parameters~\citep{shinn2023reflexionlanguageagentsverbal,zhao2024expelllmagentsexperiential,ouyang2026reasoningbankscalingagentselfevolving}.

Choosing which lessons to retain raises two connected problems. First, textual plausibility and successful retrieval do not establish a memory's decision value: a lesson may repeat what the policy already knows or reinforce an inappropriate action. Aggregate task success also leaves the contribution of an individual memory unresolved~\citep{tan2025membenchcomprehensiveevaluationmemory}. Second, continual writing increases storage, context, and maintenance costs, so retaining ineffective lessons makes future reuse more expensive~\citep{fang2026lightmemlightweightefficientmemoryaugmented}. Recent methods use action changes and trajectory feedback to guide selection~\citep{zhang2026adamemtesttimeadaptivememory,wang2026memguardpersistingverifiersignals}. We seek a candidate-level test that connects retention to behavioral evidence before a memory enters the bank:
\begin{anchorquestion}
\papericon{rq1}{lightbulb}\ \textit{Can a memory's source decision provide a practical test for whether to retain it?}
\end{anchorquestion}

We let the agent localize consequential decisions in completed trajectories and extract a memory from each. Replaying each source decision with and without its candidate under the same context reveals how that memory changes the policy (Section~\ref{sec:observation}).

This replay exposes two properties of self-extracted memory. \emph{Memory influence is concentrated:} most candidates barely shift the probed distribution, while a smaller subset produces substantial changes (Figure~\ref{fig:shift-distribution}). \emph{Memory influence is directional:} both successful and failed source actions can be reinforced or suppressed after memory injection (Figure~\ref{fig:top1-change}). A large shift alone therefore does not identify whether the effect agrees with the source role. Conversely, a success memory can strengthen an action without changing the top choice. These observations motivate testing both the magnitude of a memory's influence and its source-role alignment.

\introhead{Our approach.}
We introduce \methodname{}, which tests candidates before retention. \textbf{Localizer} extracts lessons from source decisions; \textbf{Amplituder} measures JS shift; \textbf{Aspecter} checks source-role alignment; and \textbf{BankMaster} stores, retrieves, and maintains admitted cards. Selected memories guide later trajectories, whose feedback supplies new candidates. This feedback loop between behavior and memory is the co-evolution studied here; model parameters remain frozen. Our analysis connects shift magnitude and direction to local action value under explicit conditions.

\introhead{Empirical findings.}
On ALFWorld~\citep{shridhar2021alfworldaligningtextembodied} and WebShop~\citep{yao2023webshopscalablerealworldweb}, \methodname{} ranks first across two frozen backbones. With Qwen3.5-27B, it reaches 96.27\% ALFWorld success and 73.50 WebShop score, versus 92.54\% and 62.74 for the strongest respective baselines. Source-role filtering improves ALFWorld success by 28.6\% relative to magnitude-only admission, with 41\% fewer ALFWorld cards than unfiltered retention. On WebShop with Qwen3.5-27B, token use is below all four cost-comparison baselines: slightly below MEM1 and approximately 34\%, 59\%, and 98.5\% below ReasoningBank, Reflexion, and AdaMEM, respectively.

\section{Related Work}
\label{sec:related}

\runinhead{Learning from interaction.}
Reflexion and ExpeL extract verbal lessons, Voyager accumulates skills, and CLIN maintains causal abstractions without parameter updates~\citep{shinn2023reflexionlanguageagentsverbal,zhao2024expelllmagentsexperiential,wang2023voyageropenendedembodiedagent,majumder2023clincontinuallylearninglanguage}. ReasoningBank distills strategies, while Evo-Memory, MemEvolve, and The Trinity study learning and capability evolution across task streams~\citep{ouyang2026reasoningbankscalingagentselfevolving,wei2026evomemorybenchmarkingllmagent,zhang2025memevolvemetaevolutionagentmemory,tang2026trinitycapabilitydrivenrecursiveselfimprovement}. AdaMEM uses trajectory success and action changes to select strategy-training samples; MemRL learns memory utilities from feedback~\citep{zhang2026adamemtesttimeadaptivememory,zhang2026memrlselfevolvingagentsruntime}. \methodname{} instead tests individual candidates at their source decisions before storage.

\runinhead{Memory organization and admission.}
Maintaining compact state representations over long horizons is central across visual perception and generation~\citep{tang2026rds,chen2026showdonttellmorphing} and multi-agent simulation~\citep{zhang2026gas3comprehensivesocialnetwork,zhang2026couplingmacrodynamicsmicro}; for language agents, MemGPT, Mem0, A-MEM, and SimpleMem manage, consolidate, link, or compress persistent information~\citep{packer2024memgptllmsoperatingsystems,chhikara2025mem0buildingproductionreadyai,xu2025amemagenticmemoryllm,liu2026simplememefficientlifelongmemory}. LightMem targets consolidation cost~\citep{fang2026lightmemlightweightefficientmemoryaugmented}. For selective writes, A-MAC scores candidate attributes, ConsistencyGate checks source support, and Janus compares memories on an evaluation set~\citep{zhang2026adaptivememoryadmissioncontrol,zhang2026consistencygatepreventingmemorycontamination,chen2026pastprologueplugincontroller}. MemGuard uses trajectory-verifier metadata for admission and maintenance~\citep{wang2026memguardpersistingverifiersignals}. \methodname{} adds a paired behavioral test of shift magnitude and source-role alignment.

\runinhead{Decision-aware memory.}
EPR uses model likelihoods to supervise demonstration retrieval, and CICL selects and compresses context using action-choice effects~\citep{rubin2022learningretrievepromptsincontext,guan2026decisionawarememorycardscounterfactualinspired}. Counterfactual auditing of robot memory separates behavioral influence from appropriate guidance~\citep{zhang2026memorychangesactionmemory}, echoing prefix-level preference preservation~\citep{wang2026distributionmatchingselectivemargin}. Memento-II formalizes reflective reads and writes as stateful policy improvement~\citep{wang2026memento2learningstateful}. Our focus is persistent admission through source-decision replay; the value analysis connects the resulting distribution shift to local return. Appendix~\ref{app:extended-related} provides the extended comparison.

\section{Observation: Magnitude and Direction of Memory Effects}
\label{sec:observation}

\subsection{Action Shift}
\label{sec:observation-shift}

A fixed policy $\pi_\theta$ executes task $g$, observing $o_t$, choosing $a_t\in\mathcal A_t$, and receiving feedback $r_t$. Its trajectory and pre-action history are
\begin{equation}
\begin{aligned}
\tau&=(g,o_0,a_0,r_0,\ldots,o_{T-1},a_{T-1},r_{T-1},o_T),\\
H_t&=(g,o_0,a_0,r_0,\ldots,o_t).
\end{aligned}
\end{equation}
After task completion, extractor $E$ uses the trajectory and feedback to identify consequential steps $K(\tau)$ and extract their memories:
\begin{equation}
\mathcal M(\tau)=\{m_t:m_t=E(\tau,t),\ t\in K(\tau)\}.
\label{eq:step-memory-extraction}
\end{equation}
Each memory records reusable guidance together with its source task and state.

\begin{figure}[t]
\centering
\includegraphics[width=\linewidth]{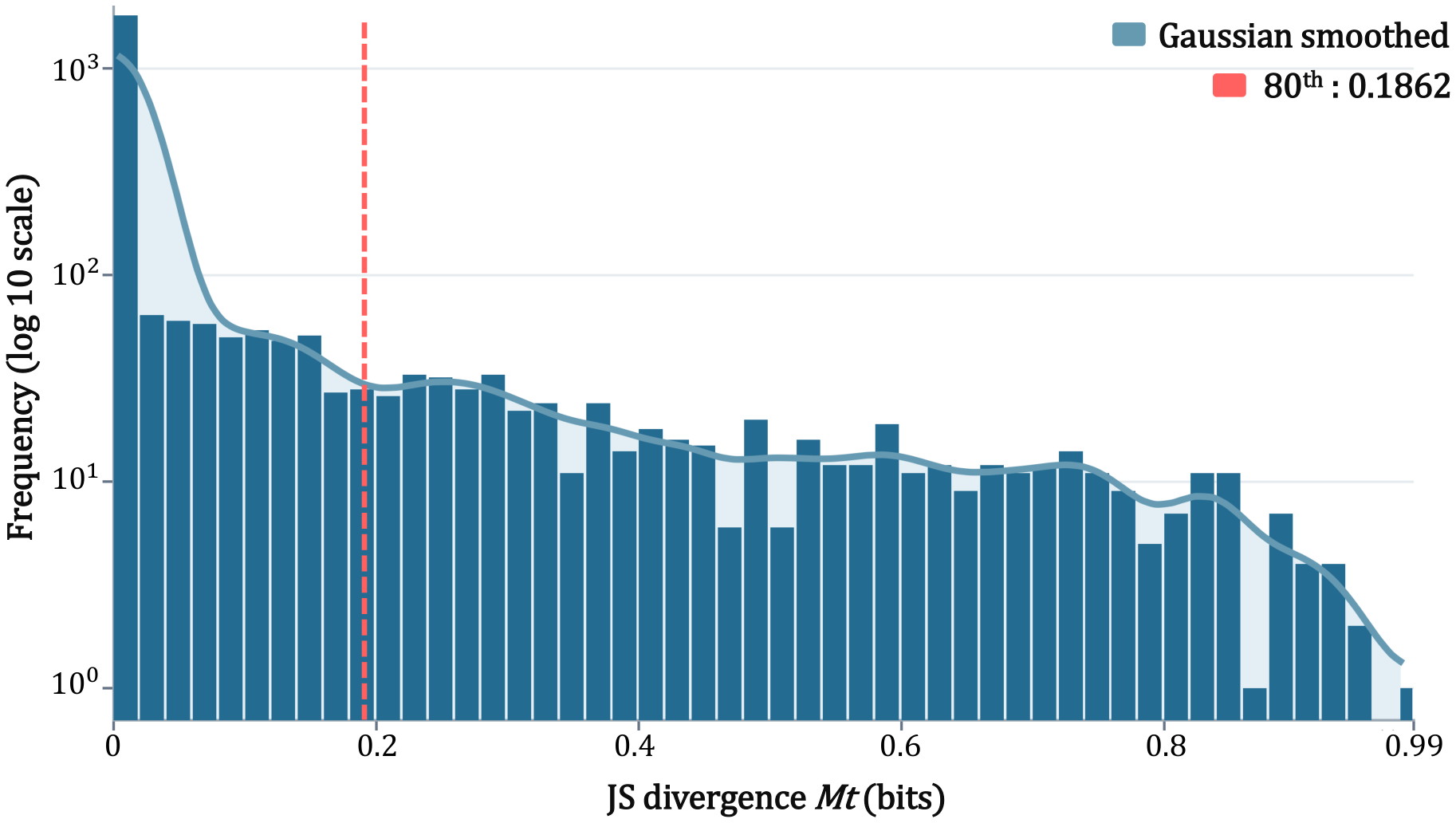}
\caption{\textbf{Shift magnitude.} Action-verb probe JS divergence for 2,785 candidate memories. Most shifts cluster near zero; a smaller tail moves the probed distribution more strongly.}
\label{fig:shift-distribution}
\end{figure}

Let $\mathcal B$ be the bank before adding these memories, $\mathcal R(\mathcal B,H_t)$ the saved source retrieval, and $F$ the context builder. The original and injected contexts are
\begin{equation}
\begin{aligned}
h_t&=F\bigl(H_t,\mathcal A_t,\mathcal R(\mathcal B,H_t)\bigr),\\
h_t^{m_t}&=F\bigl(H_t,\mathcal A_t,
\mathcal R(\mathcal B,H_t),m_t\bigr).
\end{aligned}
\label{eq:agent-memory-context}
\end{equation}
Normalize action log probabilities $\ell_t(a)$ and $\ell_t^{m_t}(a)$ from the two contexts over the same admissible set:
\begin{equation}
\begin{aligned}
P_t(a)&=\frac{\exp\ell_t(a)}{\sum_{a'\in\mathcal A_t}\exp\ell_t(a')},\\
P_t^{m_t}(a)&=\frac{\exp\ell_t^{m_t}(a)}
{\sum_{a'\in\mathcal A_t}\exp\ell_t^{m_t}(a')}.
\end{aligned}
\label{eq:observation-action-dist}
\end{equation}
We use base-2 Jensen--Shannon (JS) divergence to measure the resulting action-distribution shift:
\begin{equation}
\begin{aligned}
\bar P_t&=\tfrac12(P_t+P_t^{m_t}),\\
M_t&=D_{\mathrm{JS}}(P_t,P_t^{m_t})\\
&=\tfrac12D_{\mathrm{KL}}(P_t\|\bar P_t)
+\tfrac12D_{\mathrm{KL}}(P_t^{m_t}\|\bar P_t).
\end{aligned}
\label{eq:observation-js}
\end{equation}
Across 2,785 ALFWorld candidates, action-verb probe JS concentrates near zero, with a small tail of larger shifts (Figure~\ref{fig:shift-distribution}). Most candidates barely change action-type preference. This concentration motivates \textbf{Amplituder} (Section~\ref{sec:method-amplituder}): screen candidates by replay shift to retain memories with a measurable influence on the policy.

\subsection{Direction}
\label{sec:observation-direction}

\begin{figure}[t]
\centering
\includegraphics[width=\linewidth]{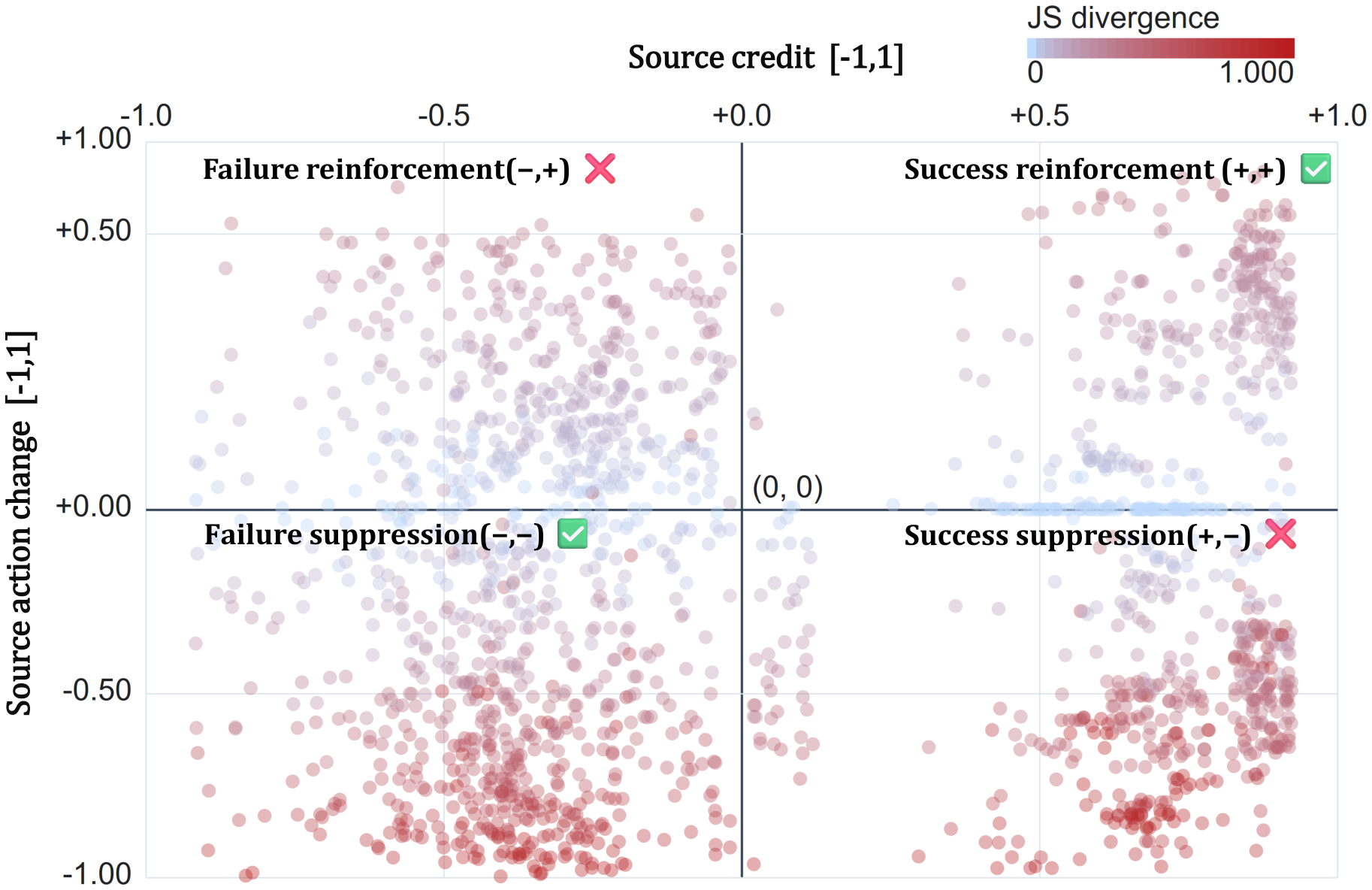}
\caption{\textbf{Shift direction.} Display-normalized source credit $c_t$ versus source-action change $\Delta p_t^{(1)}$, preserving signs. \emph{Role-aligned:} success reinforcement $(+,+)$ and failure suppression $(-,-)$. \emph{Misaligned:} success suppression $(+,-)$ and failure reinforcement $(-,+)$.}
\label{fig:top1-change}
\end{figure}

Magnitude leaves open which actions gain probability. Localizer assigns hindsight credit $c_t\in[-1,1]$ and role $y_t=\operatorname{sign}(c_t)\in\{+1,-1\}$: success maintain or failure correction. We track the recorded source action:
\begin{equation}
\begin{aligned}
a_t^{(1)}&:=a_t,\\
\Delta p_t^{(1)}&=P_t^{m_t}(a_t^{(1)})-P_t(a_t^{(1)}).
\end{aligned}
\label{eq:observation-direction}
\end{equation}
Positive $\Delta p_t^{(1)}$ reinforces the source action; negative values suppress it. Under the same ALFWorld setup, Figure~\ref{fig:top1-change} exhibits all four sign combinations.

The pairs $(c_t,\Delta p_t^{(1)})$ distinguish success reinforcement $(+,+)$, success suppression $(+,-)$, failure suppression $(-,-)$, and failure reinforcement $(-,+)$. The test $y_t\Delta p_t^{(1)}>0$ selects the two role-aligned cases. Effects in all four directions motivate \textbf{Aspecter} (Section~\ref{sec:method-aspecter}): check source-role alignment alongside magnitude, using probability changes and rankings to distinguish support for successful actions from correction of failed ones. The plot tracks source-action verbs ($L=1$).

\subsection{A Two-Axis View of Memory Effects}
\label{sec:theory}

We retain $P_t$, $P_t^{m_t}$, and $M_t$ from Section~\ref{sec:observation-shift} and analyze their complete-action form on $\mathcal A_t$. Define the displacement $v_t(a)=P_t^{m_t}(a)-P_t(a)$. Whereas Memento-II formalizes memory reads and writes as stateful policy improvement~\citep{wang2026memento2learningstateful}, our analysis relates magnitude and direction to the value of changing context at one decision.

\runinhead{Source-action consistency.}
The bounds track the recorded action $a_t^{(1)}=a_t$. Interpreting it as the baseline top choice additionally assumes $a_t=\arg\max_{a\in\mathcal A_t}\ell_t(a)$ at the complete-action level; token-wise greedy decoding alone does not establish this equality.

\runinhead{Magnitude controls effect capacity.}
Let $Q_t(a)$ be the expected remaining return after action $a$ under a fixed continuation policy, and write $\mathbb E_R[Q_t]=\sum_a R(a)Q_t(a)$. The change in expected Q-value and the action-value range are
\begin{equation}
\begin{gathered}
\mathbb E_{P_t^{m_t}}[Q_t]-\mathbb E_{P_t}[Q_t]
=\sum_a v_t(a)Q_t(a),\\
B_t=\max_{a\in\mathcal A_t}Q_t(a)-\min_{a\in\mathcal A_t}Q_t(a).
\end{gathered}
\label{eq:local-value-effect}
\end{equation}

\begin{proposition}[Radius bound]
\label{prop:magnitude-bound}
Let $\alpha_t=\operatorname{TV}(P_t,P_t^{m_t})$. For the base-2 JS divergence $M_t$,
\begin{equation}
\begin{aligned}
\left|\sum_a v_t(a)Q_t(a)\right|&\leq B_t\alpha_t\\
&\leq B_t\sqrt{2\ln2\,M_t}.
\end{aligned}
\label{eq:magnitude-bound}
\end{equation}
\end{proposition}

Near-zero complete-action divergence limits the possible expected-value change; the direction of mass transfer determines its sign.

\runinhead{Direction determines the sign.}
Write $v_t$ as probability mass transported from actions that lose mass to actions that gain it:
\begin{equation}
\sum_a v_t(a)Q_t(a)=\sum_{a,b}\mu_t(a,b)[Q_t(b)-Q_t(a)],
\label{eq:transport-value}
\end{equation}
where $\mu_t(a,b)$ transports mass from $a$ to $b$, with total mass $\alpha_t$ (Appendix~\ref{app:proof-transport}). Equal JS divergences can yield opposite value changes, depending on where this mass moves.

Using the source action $a_t^{(1)}$ and role $y_t$ from Section~\ref{sec:observation-direction}, define the signed source-role margin
\begin{equation}
\eta_t=y_t\Delta p_t^{(1)}=y_t v_t(a_t^{(1)}).
\label{eq:signed-margin}
\end{equation}
Positive $\eta_t$ corresponds to success reinforcement or failure suppression.

\begin{assumption}[Value alignment]
\label{ass:value-alignment}
The source-role margin satisfies $\eta_t\geq0$. There exists $\kappa_t\geq0$ such that every positive-mass transfer into $a_t^{(1)}$ when $y_t=+1$, or out of $a_t^{(1)}$ when $y_t=-1$, satisfies $Q_t(b)-Q_t(a)\geq\kappa_t$.
\end{assumption}
\begin{proposition}[Directional value bound]
\label{prop:direction-bound}
Under Assumption~\ref{ass:value-alignment}, the change in expected Q-value satisfies
\begin{equation}
\sum_a v_t(a)Q_t(a)\geq
\underbrace{\kappa_t\eta_t}_{\text{aligned gain}}
-\underbrace{B_t(\alpha_t-\eta_t)}_{\text{remaining loss bound}}.
\label{eq:direction-bound}
\end{equation}
Memory injection increases the expected Q-value whenever $\kappa_t\eta_t>B_t(\alpha_t-\eta_t)$.
\end{proposition}

The source-role-aligned transfers have mass $\eta_t$ and gain at least $\kappa_t\eta_t$ in value; the remaining mass $\alpha_t-\eta_t$ loses at most $B_t$ per unit. Thus, local value improves when the aligned gain exceeds the remaining loss bound. Appendix~\ref{app:proof-direction} gives the proof.

These bounds explain magnitude and direction for complete-action distributions. The implementation uses prefix likelihoods and hindsight roles as admission signals (Section~\ref{sec:method}); it tests probe response, not the unobserved Q-value conditions. Held-out task performance evaluates the retained memories' utility.

\runinhead{Transfer to future decisions.}
Let $P=(P_t)$ and $P^{\mathrm{mem}}=(P_t^{\mathrm{mem}})$ denote the baseline and memory-conditioned policies over $T$ decisions, under the same initial distribution and environment dynamics. Here $P_t^{\mathrm{mem}}(\cdot\mid H)$ includes memories retrieved at history $H$, extending the single-memory distribution $P_t^{m_t}$. With $Q_t^P(H,a)$ denoting the remaining return after action $a$ followed by $P$, extend the displacement to each history:
\begin{equation}
v_t(H,a)=P_t^{\mathrm{mem}}(a\mid H)-P_t(a\mid H).
\label{eq:retrieval-local-value}
\end{equation}
Writing $J$ for expected task return, the performance-difference identity~\citep{schulman2017trustregionpolicyoptimization} gives
\begin{equation}
\begin{aligned}
&J(P^{\mathrm{mem}})-J(P)\\
&=\mathbb E_{\tau\sim P^{\mathrm{mem}}}\!\left[
\sum_{t=0}^{T-1}\sum_a v_t(H_t,a)Q_t^P(H_t,a)\right].
\end{aligned}
\label{eq:local-to-return}
\end{equation}
When the value-alignment condition yields a nonnegative lower bound in Eq.~\ref{eq:direction-bound} at every history reached by $P^{\mathrm{mem}}$, it follows that $J(P^{\mathrm{mem}})\geq J(P)$ (Appendix~\ref{app:local-to-return}).

\begin{figure*}[t]
\centering
% Exported from figures/main.pptx; retain the editable PowerPoint as the source.
\includegraphics[width=\textwidth]{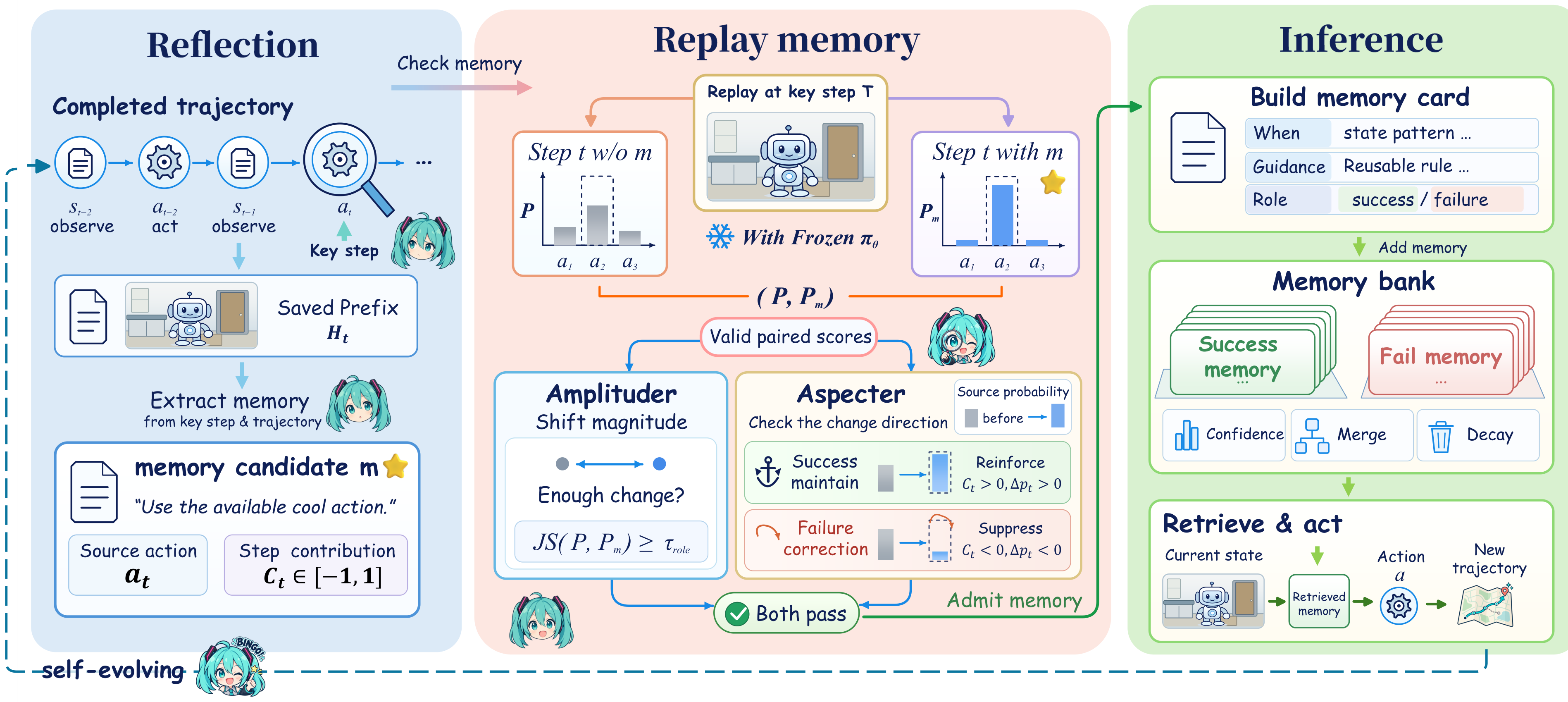}
\caption{\textbf{\methodname{} overview.} \textbf{Localizer} extracts a fixed candidate from a source decision. \textbf{Amplituder} and \textbf{Aspecter} test its shift magnitude and source-role alignment. \textbf{BankMaster} stores admitted memories and retrieves them on later tasks. Kitchen actions and probability bars are schematic.}
\label{fig:overview}
\end{figure*}

\section{Methodology}
\label{sec:method}

\subsection{\methodname: Write, Test, and Reuse}
\label{sec:method-overview}

Following Sections~\ref{sec:observation-shift}--\ref{sec:observation-direction}, \methodname{} tests \emph{how much} a memory changes the policy and \emph{whether the change supports its source role}. With $\pi_\theta$ frozen, Localizer extracts candidates, Amplituder and Aspecter test them, and BankMaster stores and reuses admitted cards (Figure~\ref{fig:overview}). Let $\mathcal B_i$ precede construction trajectory $i$. Admission decisions $G_t\in\{0,1\}$ give the update
\begin{equation}
\begin{aligned}
\mathcal M_{\mathrm{keep}}(\tau)&=\{m_t\in\mathcal M(\tau):G_t=1\},\\
\mathcal B_{i+1}&=\bankmaster(\mathcal B_i,\mathcal M_{\mathrm{keep}}(\tau),\tau).
\end{aligned}
\label{eq:method-pipeline}
\end{equation}

\subsection{Localizer: Memory Extraction}
\label{sec:method-localizer}

Localizer reflects on the completed trajectory and feedback, selecting consequential source steps with $K=5$:
\begin{equation}
K(\tau)\subseteq\{0,\ldots,T-1\},\qquad |K(\tau)|\leq K.
\label{eq:localizer-set}
\end{equation} Each hint $m_t$ receives model-assigned credit $c_t\in[-1,1]$, whose sign $y_t$ identifies its source role; zero-credit candidates are excluded. Roles describe local decisions, not episode outcomes. Importance ranks candidates; hints remain fixed during replay and become cards only after admission.

\subsection{Amplituder: Shift Magnitude}
\label{sec:method-amplituder}

Section~\ref{sec:observation-shift} shows that most candidates barely change the probed distribution. Amplituder therefore filters out negligible shifts before storage. It queries $h_t$ and $h_t^{m_t}$ without executing an environment action, preserving the saved history, task guidance, retrieved cards, and candidate actions. Only the candidate-hint block differs; the resulting distributions yield $M_t$ in Eq.~\ref{eq:observation-js}.

\runinhead{Token-budgeted scoring.}
We sum log probabilities of the first $L$ action-content tokens after \texttt{<action>} and normalize over distinct prefixes, yielding $\widetilde P_t$ and $\widetilde P_t^{m_t}$. ALFWorld uses $L=1$; WebShop uses $L=5$. Shared prefixes form one category; admission tracks the source category $v_t^{(1)}=\phi_L(a_t^{(1)})$. Candidate construction, normalization, and complete-action prompts are detailed in Appendices~\ref{app:prototype-delta} and~\ref{app:prompts}.

Both roles must cross a magnitude threshold:
\begin{equation}
\begin{aligned}
A_t^+&\Longleftrightarrow M_t\geq\tau_{\mathrm{succ}},\\
A_t^-&\Longleftrightarrow M_t\geq\tau_{\mathrm{fail}}.
\end{aligned}
\label{eq:amplituder-score}
\end{equation}
We use $0.05$ and $0.12$ bits, respectively. The lower success threshold accommodates source actions already strongly preferred before injection.

\subsection{Aspecter: Source-Role Alignment}
\label{sec:method-aspecter}

Section~\ref{sec:observation-direction} motivates checking whether memory supports a successful source decision or corrects a failed one. Recall that $a_t^{(1)}$ is the recorded source action and $\Delta p_t^{(1)}$ is its probability change after injection (Eq.~\ref{eq:observation-direction}). The top choices before and after injection are
\begin{equation}
\begin{aligned}
\widehat a_t^{0}&=\arg\max_{a\in\mathcal A_t}P_t(a),\\
\widehat a_t^{m_t}&=\arg\max_{a\in\mathcal A_t}P_t^{m_t}(a).
\end{aligned}
\label{eq:aspecter-score}
\end{equation}
The predicates below use complete-action notation; the implementation substitutes $\widetilde P_t$, $\widetilde P_t^{m_t}$, and source category $v_t^{(1)}$ on the fixed prefix set (Appendix~\ref{app:algorithm-predicates}).

\begin{subequations}
\label{eq:aspecter-branches}
\Needspace{8\baselineskip}
\subsubsection{Success Maintain}
The reinforcement condition $S_t^{\mathrm{gain}}$ requires the injected top choice to equal the source action and its probability to increase by at least $\epsilon_{\mathrm{succ}}$:
\begin{equation}
\begin{aligned}
S_t^{\mathrm{gain}}\Longleftrightarrow{}&
(\widehat a_t^{m_t}=a_t^{(1)})\\
&\land(\Delta p_t^{(1)}\geq\epsilon_{\mathrm{succ}}).
\end{aligned}
\label{eq:aspecter-gain}
\end{equation}
\subsubsection{Failure Corrections}
The correction condition $F_t^{\mathrm{dir}}$ accepts either source suppression by at least $\epsilon_{\mathrm{drop}}$ or an injected preference for an alternative:
\begin{equation}
\begin{aligned}
F_t^{\mathrm{dir}}\Longleftrightarrow{}&
(\Delta p_t^{(1)}\leq-\epsilon_{\mathrm{drop}})\\
&\lor(\widehat a_t^{m_t}\neq a_t^{(1)}).
\end{aligned}
\label{eq:aspecter-failure}
\end{equation}
\end{subequations}
The ranking branch tests alternative preference, which can hold even without a source-probability decrease; it is distinct from positive signed-margin alignment in Eq.~\ref{eq:signed-margin}.

\noindent\textbf{Combined admission.}
Let $V_t$ denote comparable probe scores and $A_t^+,A_t^-$ the magnitude conditions (Eq.~\ref{eq:amplituder-score}). Admission follows the source role:
\begin{equation}
\begin{aligned}
y_t=+1:\quad G_t=1\Longleftrightarrow{}&
V_t\land A_t^+\land S_t^{\mathrm{gain}},\\
y_t=-1:\quad G_t=1\Longleftrightarrow{}&
V_t\land A_t^-\land F_t^{\mathrm{dir}}.
\end{aligned}
\label{eq:method-admission}
\end{equation}
Otherwise $G_t=0$. Appendix~\ref{app:implementation-parameters} lists threshold values; Algorithm~\ref{alg:branch-admission} implements the gate.

\subsection{BankMaster: Retrieval and Maintenance}
\label{sec:method-bankmaster}

BankMaster stores each admitted hint verbatim as card guidance, adding only role, applicability, retrieval keys, and source metadata. Card construction preserves the tested text. It retrieves the highest-scoring compatible cards:
\begin{equation}
\begin{aligned}
\mathcal{B}_{i}^R(H_t)
&=\{m\in\mathcal{B}_{i}:
s_{\mathrm{ret}}(m\mid H_t,\mathcal A_t)\geq\tau_R\},\\
\mathcal R(\mathcal{B}_{i},H_t)
&=\operatorname{TopK}_{m\in\mathcal{B}_{i}^R(H_t)}
s_{\mathrm{ret}}(m\mid H_t,\mathcal A_t).
\end{aligned}
\label{eq:bankmaster-retrieval}
\end{equation}
Here $s_{\mathrm{ret}}$ combines task-family and state similarity with action compatibility. Insertion and feedback-driven maintenance update the bank:
\begin{equation}
\begin{aligned}
\mathcal{B}_{i}^+&=\operatorname{Insert}\bigl(
\mathcal{B}_{i},\{\operatorname{Card}(m_t):G_t=1\}\bigr),\\
\mathcal{B}_{i+1}&=\operatorname{Maintain}\bigl(
\mathcal{B}_{i}^+,\operatorname{Feedback}(\tau)\bigr).
\end{aligned}
\label{eq:bankmaster-update}
\end{equation}
Maintenance updates confidence, merges similar same-role cards, decays stale records, and prunes at capacity. Appendix~\ref{app:bankmaster-details} provides details. Held-out evaluation uses read-only retrieval.

\begin{table*}[!t]
\centering
\color{black}
\footnotesize
\setlength{\tabcolsep}{3.5pt}
\renewcommand{\arraystretch}{1.04}
\setlength{\extrarowheight}{0pt}
\anchortablestyle
\caption{\rqmark{1}\ \textbf{Held-out task performance.} ALFWorld success (\%) and WebShop mean score and steps. Bold and underline mark the best and second-best results per backbone; task-family ties use Overall.}
\label{tab:main-results}
\begin{tabular*}{\textwidth}{@{\extracolsep{\fill}}l*{9}{r}@{}}
\toprule
& \multicolumn{7}{c}{\tableheader{ALFWorld}} & \multicolumn{2}{c}{\tableheader{WebShop}}\\
\cmidrule(lr){2-8}\cmidrule(lr){9-10}
\addlinespace[5pt]
\tableheader{Method} & \tableheader{Overall}\,$\uparrow$ & \tableheader{Put}\,$\uparrow$ & \tableheader{Clean}\,$\uparrow$ & \tableheader{Heat}\,$\uparrow$ & \tableheader{Cool}\,$\uparrow$ & \tableheader{Examine}\,$\uparrow$ & \tableheader{Puttwo}\,$\uparrow$ & \tableheader{Score}\,$\uparrow$ & \tableheader{Steps}\,$\downarrow$\\
\addlinespace[5pt]
\midrule
\multicolumn{10}{c}{\modelgroup{Qwen2.5-14B-Instruct}}\\
\midrule
ReAct & 46.27 & 33.33 & 51.61 & 26.09 & \second{85.71} & 72.22 & 5.88 & 41.77 & 8.68\\
\addlinespace[2.5pt]
\multicolumn{10}{l}{\tablegroup{layer-group}{Conventional memory methods}}\\
Mem0 & 12.69 & 0.00 & 0.00 & 4.35 & 0.00 & \second{88.89} & 0.00 & 38.32 & 9.11\\
Reflexion & 67.16 & \second{77.78} & 55.91 & 66.67 & 74.60 & 77.78 & 52.94 & 25.24 & 8.41\\
MemP & 56.72 & 63.89 & 63.44 & 52.17 & 50.79 & 72.22 & 31.37 & 44.61 & 16.21\\
LangMem & 36.57 & 12.50 & 32.26 & 26.09 & 71.43 & 83.33 & 0.00 & 36.60 & 10.12\\
SimpleMem & 32.84 & 23.61 & 12.90 & 46.38 & 30.16 & 62.96 & 35.29 & 45.46 & 7.31\\
ExpeL & 62.69 & 41.67 & \second{70.97} & \second{73.91} & 80.95 & 72.22 & 29.41 & 53.92 & 8.71\\
\addlinespace[2.5pt]
\multicolumn{10}{l}{\tablegroup{sync-alt}{Self-evolving memory methods}}\\
MEM1 & 30.35 & 23.61 & 23.66 & 23.19 & 28.57 & 66.67 & 25.49 & 34.46 & 9.95\\
AdaMEM & \second{72.64} & 73.61 & 69.89 & 69.57 & 74.60 & 68.52 & \best{82.35} & 36.75 & 9.96\\
ReasoningBank & 53.48 & 45.83 & 56.99 & 46.38 & 58.73 & 55.56 & 58.82 & \second{59.05} & \best{4.84}\\
\midrule
\anchorrowname & \best{87.06} & \best{91.67} & \best{88.17} & \best{82.61} & \best{88.89} & \best{92.59} & \second{76.47} & \best{65.88} & \second{6.32}\\
\midrule
\multicolumn{10}{c}{\modelgroup{Qwen3.5-27B}}\\
\midrule
ReAct & 61.19 & 66.67 & 51.61 & 65.22 & 76.19 & 72.22 & 35.29 & 45.97 & 10.64\\
\addlinespace[2.5pt]
\multicolumn{10}{l}{\tablegroup{layer-group}{Conventional memory methods}}\\
Mem0 & 11.94 & 0.00 & 0.00 & 0.00 & 0.00 & 88.89 & 0.00 & 52.02 & 9.18\\
Reflexion & 91.79 & \second{100.00} & \second{90.32} & \best{95.65} & 76.19 & 88.89 & \second{100.00} & 48.18 & 17.32\\
MemP & 57.71 & 98.61 & 81.72 & 1.45 & 19.05 & 61.11 & 76.47 & 59.83 & 14.51\\
LangMem & 59.70 & 47.22 & 53.76 & 44.93 & 44.44 & 100.00 & 84.31 & 54.85 & 8.90\\
SimpleMem & \second{92.54} & 95.83 & 89.25 & 82.61 & \second{95.24} & \second{100.00} & 96.08 & 61.68 & 7.19\\
ExpeL & 82.84 & 75.00 & 90.32 & 73.91 & 95.24 & 77.78 & 82.35 & 55.44 & 8.59\\
\addlinespace[2.5pt]
\multicolumn{10}{l}{\tablegroup{sync-alt}{Self-evolving memory methods}}\\
MEM1 & 57.46 & 50.00 & 52.69 & 43.48 & 41.27 & 100.00 & 70.59 & 48.28 & 9.47\\
AdaMEM & 65.67 & 98.61 & 33.33 & 17.39 & 76.19 & 100.00 & 94.12 & 59.25 & 49.00\\
ReasoningBank & 80.35 & 79.17 & 77.42 & 76.81 & 57.14 & 100.00 & 100.00 & \second{62.74} & \second{6.61}\\
\midrule
\anchorrowname & \best{96.27} & \best{100.00} & \best{90.32} & \second{91.30} & \best{100.00} & \best{100.00} & \best{100.00} & \best{73.50} & \best{5.33}\\
\bottomrule
\end{tabular*}
\end{table*}

\Needspace{6\baselineskip}
\section{Experiments}
\label{sec:experiments}

Our experiments address four research questions:

\runinhead{\rqmark{1}: Memory utility.}
Can the extracted memories improve held-out task performance across environments and backbones while retaining fewer memories than storing every candidate?

\runinhead{\rqmark{2}: Admission design.}
How do shift magnitude and source-role direction contribute to memory quality, and what is gained by combining success memories with failure corrections?

\runinhead{\rqmark{3}: Resource efficiency.}
How does \methodname{} balance task performance, persistent memory size, and total token use across memory construction and evaluation?

\runinhead{\rqmark{4}: Self-evolution.}
As experience accumulates along a task stream, does selective memory reuse improve the frozen agent's performance beyond simply expanding its memory bank?

\subsection{Experimental Settings}
\label{sec:experimental-settings}

\runinhead{Backbones and baselines.}
We freeze either Qwen2.5-14B-Instruct~\citep{qwenteam2024qwen25} or Qwen3.5-27B~\citep{qwenteam2026qwen35} and compare ReAct~\citep{yao2022react}, Mem0~\citep{chhikara2025mem0buildingproductionreadyai}, Reflexion~\citep{shinn2023reflexionlanguageagentsverbal}, MemP~\citep{fang2026mempexploringagentprocedural}, LangMem~\citep{langchain2025langmem}, SimpleMem~\citep{liu2026simplememefficientlifelongmemory}, ExpeL~\citep{zhao2024expelllmagentsexperiential}, MEM1~\citep{zhou2025mem1learningsynergizememory}, AdaMEM~\citep{zhang2026adamemtesttimeadaptivememory}, and ReasoningBank~\citep{ouyang2026reasoningbankscalingagentselfevolving}. All methods share the evaluation split within each backbone. We use each baseline's open-source benchmark implementation when available; otherwise, its memory mechanism uses the same environment and action-selection prompt as \methodname{}. Appendix~\ref{app:prototype-delta} details adaptation.

\runinhead{Tasks and metrics.}
Memory construction uses 200 ALFWorld tasks and 400 WebShop tasks. Held-out evaluation uses 402 ALFWorld tasks and 4,055 WebShop instructions from a 10,000-product subset. ALFWorld reports success overall and by family; Overall pools successes across all tasks. WebShop reports mean task score and interaction steps. Evaluation is read-only: outcomes do not update the constructed banks. Appendix~\ref{app:implementation-parameters} gives task counts and implementation settings.

\subsection{Main Results: Performance on Held-Out Tasks}
\label{sec:main-results}

\begin{figure*}[t]
\centering
\includegraphics[width=\textwidth]{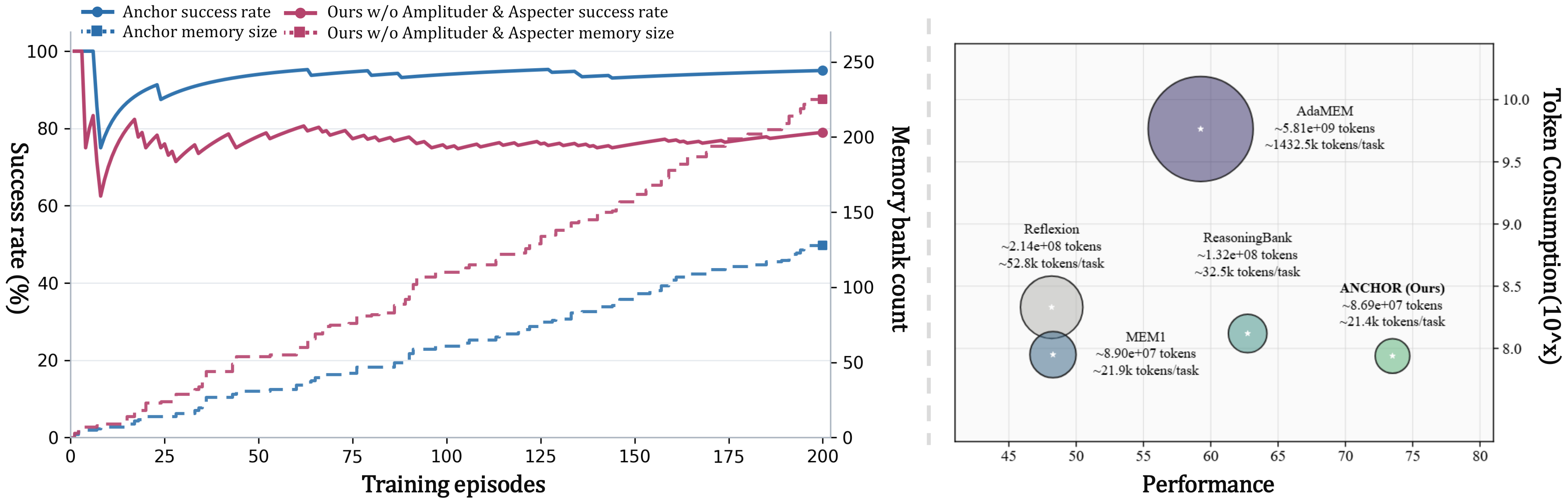}
\caption{\rqmark{3} and \rqmark{4}\ \textbf{Success, bank size, and token consumption with Qwen3.5-27B.} (\textbf{\textit{Left}}) Sample success rate (cumulative; solid) over 200 randomly sampled ALFWorld memory-construction tasks and bank size (dashed) for \methodname{} and \ablnobehavior{}. (\textbf{\textit{Right}}) WebShop mean score versus total input and output tokens across construction and evaluation ($10^x$ scale). Circle area is proportional to mean interaction steps; lower, smaller circles indicate lower cost.}
\label{fig:memory-efficiency}
\end{figure*}

Table~\ref{tab:main-results} compares held-out performance across two environments, two frozen backbones, and ten baselines. Table~\ref{tab:admission-ablation} and Figure~\ref{fig:memory-efficiency} connect these gains to selective retention and bank size.

\resultobs{1}{Selected memories improve held-out performance with a smaller bank.}
\methodname{} ranks first in all four environment--backbone combinations. On ALFWorld, it retains 41\% fewer memories than writing every candidate while raising held-out success from 73.38\% to 96.27\% (Table~\ref{tab:admission-ablation}; Figure~\ref{fig:memory-efficiency}). Thus, selective retention improves the usefulness of accumulated experience, not just its storage footprint.

\runinhead{ALFWorld task success.}
With Qwen2.5-14B, \methodname{} reaches 87.06\% success, exceeding AdaMEM (72.64\%) by 14.43 points (19.9\% relatively). With Qwen3.5-27B, it reaches 96.27\%, surpassing SimpleMem (92.54\%) by 3.73 points. Differences are calculated before rounding. Gains span task families: \methodname{} leads five of six with Qwen2.5 and reaches at least 90.32\% in every family with Qwen3.5.

\Needspace{7\baselineskip}
\runinhead{WebShop score and interaction efficiency.}
\methodname{} scores 65.88 and 73.50 versus ReasoningBank's 59.05 and 62.74, improving by 11.6\% and 17.2\% relatively. It also reduces Qwen3.5 mean steps from 6.61 to 5.33 and ranks second in Qwen2.5 step efficiency. Retained guidance improves outcomes without longer interactions.

\begin{table}[!t]
\centering
\small
\setlength{\tabcolsep}{3pt}
\renewcommand{\arraystretch}{1.08}
\anchortablestyle
\caption{\rqmark{2}\ \textbf{Overall ALFWorld admission ablation.} Qwen3.5-27B success (\%). Underline marks the best ablation.}
\label{tab:admission-ablation}
\begin{tabular}{>{\raggedright\arraybackslash}p{\dimexpr.70\columnwidth-2\tabcolsep\relax}>{\raggedleft\arraybackslash}p{\dimexpr.30\columnwidth-2\tabcolsep\relax}}
\toprule
\tableheader{Admission variant} & \tableheader{Success (\%)}\\
\midrule
\tablemark{\papericon{anchorInk}{robot}}ReAct (no memory) & 61.19\\
\tablemark{\papericon{anchorInk}{layer-group}}\ablnobehavior{} & 73.38\\
\addlinespace[2pt]
\tablemark{\amplitudermark}\ablnoaspecter{} & 74.88\\
\tablemark{\successmark}\ablnofailure{} & \second{86.82}\\
\tablemark{\failuremark}\ablnosuccess{} & 83.83\\
\midrule
\anchorrowname & \best{96.27}\\
\bottomrule
\end{tabular}
\end{table}

\Needspace{7\baselineskip}
\subsection{Contributions of Magnitude and Direction}
\label{sec:filter-analysis}

Table~\ref{tab:admission-ablation} uses Qwen3.5-27B on the same ALFWorld split. \ablnobehavior{} writes every candidate; \ablnoaspecter{} uses magnitude alone. Role ablations keep one role with its tests intact. Appendix Table~\ref{tab:admission-ablation-full} gives family-level results.

\resultobs{2}{Direction improves magnitude-only admission; retaining both roles performs best.}
Removing both tests yields 73.38\%; magnitude filtering alone reaches 74.88\%, a 1.49-point difference. Adding direction raises success to 96.27\%, a further 21.39 points (28.6\% relative). The larger gain comes from distinguishing source-role effects among candidates that pass magnitude screening. Keeping both roles adds 9.45 points over success memories alone and 12.44 over failure corrections alone. Thus, the strongest tested variant combines behavioral screening with both reinforcement and correction memories.

\subsection{Resource Use}
\label{sec:efficiency-analysis}

\resultobs{3}{Selective retention improves the performance--resource trade-off.}
In Figure~\ref{fig:memory-efficiency}, \methodname{} stores 41\% fewer ALFWorld memories while achieving higher success. On WebShop with Qwen3.5-27B, it leads the five plotted methods in score, mean steps, and token use (approximately 86.9M). Token use is slightly below MEM1 and approximately 34\%, 59\%, and 98.5\% below ReasoningBank, Reflexion, and AdaMEM, respectively. These comparisons measure bank size, token consumption, and interaction steps. Token totals cover construction and evaluation, including reflection, replay, and maintenance (Appendix~\ref{app:prototype-delta}).

\Needspace{8\baselineskip}
\subsection{Self-Evolution Along the Task Stream}
\label{sec:online-evolution}

\resultobs{4}{Selective accumulation improves construction-stream performance and held-out reuse.}
Figure~\ref{fig:memory-efficiency} tracks cumulative success over 200 randomly sampled ALFWorld construction tasks. \methodname{} and \ablnobehavior{} share a frozen backbone but differ in admission. After early fluctuations, \methodname{} maintains higher cumulative success, finishing approximately 20\% higher relatively despite slower bank growth. This curve compares behavior during memory construction; Table~\ref{tab:admission-ablation} tests the resulting banks on held-out tasks. Together, they connect selective experience accumulation to stronger subsequent reuse without parameter updates.

\Needspace{8\baselineskip}
\section{Conclusion}
\label{sec:conclusion}

We introduced \methodname{}, which tests memories at their source decisions before retention. Paired replay measures shift magnitude and source-role alignment under a frozen policy. Across ALFWorld and WebShop, \methodname{} improves held-out performance; resource comparisons show smaller banks and lower token use. These results highlight memory curation as a practical link between reflection and behavioral improvement. By letting experience shape memory and selected memory guide subsequent behavior, \methodname{} offers a path toward experience-driven agent co-evolution.

\section*{Limitations}

The scope of \methodname{} depends on the resolution of its behavioral probe and access to model probabilities.

\runinhead{Action-space requirements.}
Our long-horizon benchmarks provide explicit candidate actions. Prefix scoring cannot distinguish actions sharing the scored tokens, including some object-level errors. Open-ended settings require other consistent measures of memory-induced change, such as plan or response differences.

\runinhead{Access to model probabilities.}
Token log probabilities are required but unavailable through some closed-source interfaces. Model-generated scores, rankings, or repeated-output preferences offer alternatives. Future work can calibrate these measures and extend replay to broader tasks and interfaces.

\bibliography{references}

\clearpage
\appendix
% Shared layout for all supplementary sections.
% Single-column appendix for the arXiv preprint.
\onecolumn
\setcounter{dbltopnumber}{3}
\renewcommand{\dblfloatpagefraction}{.88}
\renewcommand{\floatpagefraction}{.82}
\renewcommand{\topfraction}{.92}
\renewcommand{\bottomfraction}{.85}
\renewcommand{\textfraction}{.08}
\normalfont\normalsize
\setlength{\parskip}{1.8pt plus 0.8pt minus 0.4pt}
\abovedisplayskip=5.5pt plus 1.5pt minus 1.5pt
\belowdisplayskip=5.5pt plus 1.5pt minus 1.5pt
\newcommand{\appendixfigfont}{\fontsize{9.6}{11.8}\selectfont}
\newcommand{\appsuccessbadge}{\textcolor{rq3}{\faIcon{check-circle}}\;\textbf{\textcolor{rq3}{Pass}}}
\newcommand{\apprejectbadge}{\textcolor{rq4}{\faIcon{times-circle}}\;\textbf{\textcolor{rq4}{Reject}}}
\newcommand{\promptvar}[1]{\texttt{\textcolor{rq1}{\{#1\}}}}
\newcommand{\promptkey}[1]{\textbf{\textcolor{anchorInk}{#1}}}

% Refined visual hierarchy for single-column supplementary headings.
\makeatletter
\renewcommand\section{\@startsection{section}{1}{\z@}%
  {-2.2ex plus -0.5ex minus -0.2ex}%
  {1.1ex plus 0.2ex}%
  {\normalfont\large\bfseries\color{anchorInk}}}
\renewcommand\subsection{\@startsection{subsection}{2}{\z@}%
  {-1.7ex plus -0.4ex minus -0.2ex}%
  {0.7ex plus 0.2ex}%
  {\normalfont\normalsize\bfseries\color{anchorInk}}}
\renewcommand\paragraph{\@startsection{paragraph}{4}{\z@}%
  {0.9ex plus 0.3ex minus 0.1ex}%
  {-0.8em}%
  {\normalfont\normalsize\bfseries\color{anchorInk}}}
\makeatother

\tcbset{
  appendixbox/.style={enhanced,colback=white,colframe=anchorInk!24,
    boxrule=.45pt,arc=1.2mm,colbacktitle=groupgray,coltitle=anchorInk,
    fonttitle=\bfseries\appendixfigfont,
    toptitle=2.5pt,bottomtitle=2.5pt,left=7.5pt,right=7.5pt,top=4pt,bottom=4pt,
    before skip=4.5pt,after skip=4.5pt},
  appendixtoc/.style={enhanced,colback=groupgray!65,colframe=anchorInk!22,
    borderline west={2.2pt}{0pt}{rq1},sharp corners=west,arc=1.2mm,
    boxrule=.45pt,left=12pt,right=12pt,top=8pt,bottom=8pt,
    before skip=6pt,after skip=12pt},
  appendixthm/.style={enhanced,breakable,colback=rq1!3,colframe=rq1!22,
    borderline west={1.8pt}{0pt}{rq1},sharp corners=west,arc=1mm,
    boxrule=.35pt,left=9pt,right=9pt,top=5pt,bottom=5pt,
    before skip=6pt,after skip=6pt},
  appendixprompt/.style={appendixbox,breakable,
    title after break={\alfmark\hspace{.4em}\textit{Prompt template (continued)}\hfill\textit{Role: User}},
    pad at break*=2pt,
    colback=groupgray!32,colframe=rq1!26,colbacktitle=rq1!9,coltitle=anchorInk,
    borderline west={1.8pt}{0pt}{rq1!75},
    fontupper=\appendixfigfont\raggedright,
    toptitle=1.8pt,bottomtitle=1.8pt,left=7pt,right=7pt,top=2.5pt,bottom=2.5pt,
    before skip=2.8pt,after skip=2.8pt},
  appendixpromptws/.style={appendixprompt,
    title after break={\webmark\hspace{.4em}\textit{Prompt template (continued)}\hfill\textit{Role: User}},
    colframe=rq3!28,colbacktitle=rq3!9,
    borderline west={1.8pt}{0pt}{rq3!75}},
  appendixalgorithm/.style={appendixbox,
    colback=groupgray!28,colframe=anchorInk!28,colbacktitle=anchorInk!8,
    borderline west={2pt}{0pt}{anchorInk},
    fontupper=\appendixfigfont,left=9pt,right=9pt,top=5.5pt,bottom=5.5pt},
  appendixexample/.style={appendixbox,
    fontupper=\appendixfigfont\raggedright,
    fonttitle=\bfseries\appendixfigfont,
    left=7.5pt,right=7.5pt,top=3.5pt,bottom=3.5pt,before skip=4pt,after skip=4pt},
  appendixsuccess/.style={appendixexample,colback=oursgray!38,
    colbacktitle=oursgray!95,colframe=rq3!42,coltitle=rq3!70!black,
    borderline west={2pt}{0pt}{rq3}},
  appendixfailure/.style={appendixexample,colback=rq4!4,
    colbacktitle=rq4!13,colframe=rq4!42,coltitle=rq4!75!black,
    borderline west={2pt}{0pt}{rq4}},
  appendixrecord/.style={appendixexample,fontupper=\appendixfigfont,
    fonttitle=\bfseries\appendixfigfont,colback=groupgray!42,
    colbacktitle=anchorInk!8,colframe=anchorInk!26,coltitle=anchorInk,
    borderline west={1.8pt}{0pt}{anchorInk!70}}
}

\begin{tcolorbox}[appendixtoc]
{\large\bfseries\color{anchorInk}\papericon{rq1}{compass}\hspace{.45em}Appendix Overview \& Contents}\par\vspace{5pt}
\begingroup
\hypersetup{linkcolor=anchorInk}
\small
\newcommand{\appcontentssection}[3]{%
  \par\vspace{2pt}\noindent
  \hyperref[#1]{\textbf{\textcolor{rq1}{Appendix~\ref*{#1}}}\hspace{.55em}\textbf{#2}}%
  \hspace{.4em}{\color{black!55}--- #3}\dotfill\textbf{\pageref{#1}}\par}
\appcontentssection{app:extended-related}{Extended Related Work}{Interaction learning, memory organization, selective updates, and social systems}
\appcontentssection{app:proofs}{Theoretical Analysis}{Radius, transport, directional value, return identity, and finite-sample bounds}
\appcontentssection{app:experiment-details}{Experimental Details}{Family ablations, BankMaster maintenance, parameters, and token accounting}
\appcontentssection{app:algorithm}{Algorithms}{Write-and-reuse evolution loop and branch-specific admission gate}
\appcontentssection{app:prompts}{Prompt Templates}{Complete runtime prompts for ALFWorld and WebShop across all five stages}
\appcontentssection{app:cases}{Case Studies}{Paired source-state probes, memory card gallery, and inference-time traces}
\endgroup
\end{tcolorbox}

\section{Extended Related Work}
\label{app:extended-related}

\runinhead{Learning from interaction.}
Deliberate search and adaptive reasoning expand single-episode decision-making through tree search and dual-mode knowledge-assisted planning~\citep{yao2023treethoughtsdeliberateproblem,zhou2024languageagenttreesearch,chen2025finequestadaptiveknowledgeassistedsports}.
Across episodes, Reflexion stores verbal feedback, ExpeL extracts lessons, AutoGuide synthesizes state-aware guidelines, Agent Workflow Memory induces reusable routines, and Synapse reuses trajectories~\citep{shinn2023reflexionlanguageagentsverbal,zhao2024expelllmagentsexperiential,fu2024autoguideautomatedgenerationselection,wang2024agentworkflowmemory,zheng2024synapsetrajectoryasexemplarpromptingmemory}; Voyager accumulates executable skills~\citep{wang2023voyageropenendedembodiedagent}. CLIN updates persistent causal abstractions across trials without parameter updates~\citep{majumder2023clincontinuallylearninglanguage}. ReasoningBank distills strategies, LatentMorph condenses intermediate states into compact visual memory to steer token predictions, Evo-Memory evaluates learning across task streams, and MemEvolve adapts experiential knowledge and its management~\citep{ouyang2026reasoningbankscalingagentselfevolving,chen2026showdonttellmorphing,wei2026evomemorybenchmarkingllmagent,zhang2025memevolvemetaevolutionagentmemory}. AdaMEM combines long-term trajectories with within-episode strategies; its Step-MFT selects strategy-training samples using trajectory success and action changes~\citep{zhang2026adamemtesttimeadaptivememory}. MemRL learns memory utilities from execution feedback, while MACE co-evolves agents and memory graphs in multi-agent systems~\citep{zhang2026memrlselfevolvingagentsruntime,yang2026macememoryagentcoevolutionadaptive}. \methodname{} uses source decisions and feedback roles to select candidates before storage.

Mem$^2$Evolve couples experience distillation with the creation of tools and expert agents~\citep{cheng2026mem2evolveselfevolvingagentscoevolutionary}, SkillRL co-evolves a hierarchical skill bank and the policy through reinforcement learning~\citep{xia2026skillrlevolvingagentsrecursive}, and verification-grounded agent frameworks span formal theory exploration in VALG, package-scoped scientific investigation in EarthVerse, and capability-driven harness evolution in The Trinity~\citep{zhang2026valgagenticmltheory,cui2026earthversebenchmarkingscientificagents,tang2026trinitycapabilitydrivenrecursiveselfimprovement}. \methodname{} selects external lessons for retention under a fixed policy and action interface.

\runinhead{Memory organization and efficient reuse.}
Surveys characterize agent memory by its functions, representations, and evaluation~\citep{zhang2024surveymemorymechanismlarge}; CoALA connects modular memory with structured actions and decision-making~\citep{sumers2024cognitivearchitectureslanguageagents}, complementing compact recurrent state updates in long-horizon visual tracking~\citep{tang2026rds}.
\citet{liu2023lostmiddlelanguagemodels} show that information position affects performance, motivating careful selection and organization of retrieved experience.
MemGPT manages information beyond the context window, Generative Agents combines retrieval with reflection, HippoRAG indexes long-term knowledge via graph-based spreading activation, and MemoryBank adds summarization and forgetting to conversational memory~\citep{packer2024memgptllmsoperatingsystems,park2023generativeagentsinteractivesimulacra,gutierrez2025hipporagneurobiologicallyinspiredlongterm,zhong2023memorybankenhancinglargelanguage}. Mem0 consolidates facts, A-MEM links notes, and SimpleMem compresses memories~\citep{chhikara2025mem0buildingproductionreadyai,xu2025amemagenticmemoryllm,liu2026simplememefficientlifelongmemory}. LightMem addresses consolidation cost, while MemHarness reconstructs experience for the current state~\citep{fang2026lightmemlightweightefficientmemoryaugmented,wu2026memharnessmemoryreconstructedreplayed}. MemBench, LongMemEval, and MemoryAgentBench test the effectiveness of sustained memory use~\citep{tan2025membenchcomprehensiveevaluationmemory,wu2025longmemevalbenchmarkingchatassistants,hu2026evaluatingmemoryllmagents}. \methodname{} adds behavioral admission to this memory-management pipeline.

Reflective Memory Management combines multi-granularity construction and feedback-driven retrieval for dialogue~\citep{tan2025prospectretrospectreflectivememory}, while personalized interactive systems adapt task progression from ongoing user responses~\citep{shi2026letgamesllmpoweredgamifiedapproach}. \methodname{} places a paired source-decision test between extraction and storage, evaluating both construction-stream learning and read-only transfer.

\runinhead{Selective and reliable memory updates.}
Self-Refine iteratively revises outputs using model-generated feedback~\citep{madaan2023selfrefine}, whereas intrinsic self-correction often falters without external grounding~\citep{huang2024largelanguagemodelsselfcorrect}, motivating step-level process verification, tool-interactive critiquing in CRITIC, and calibrated adversarial probing in CodeHacker~\citep{lightman2023letsverifystepstep,gou2024criticlargelanguagemodels,shi2026codehackerautomatedtestcase}. For persistent memory, A-MAC scores candidate attributes, AdaMem learns preference-conditioned writes, and D-MEM routes selected inputs into a slower evolution path~\citep{zhang2026adaptivememoryadmissioncontrol,chen2026adamemlearningrememberadaptive,song2026dmemdopaminegatedagenticmemory}. ConsistencyGate checks source support, TRUSTMEM checks update properties, and Janus compares memories on a compact evaluation set~\citep{zhang2026consistencygatepreventingmemorycontamination,yang2026trustmemlearningtrustworthymemory,chen2026pastprologueplugincontroller}. MemGuard stores trajectory-verifier signals as metadata for admission and later maintenance~\citep{wang2026memguardpersistingverifiersignals}. Memory poisoning, state contamination, and persistent injection expose risks of unverified writes~\citep{chen2024agentpoisonredteamingllmagents,wang2026statecontaminationmemoryaugmentedllm,yang2026zombieagentspersistentcontrol}. \methodname{} directly measures local action-distribution changes and source-role alignment, extending admission beyond content- and outcome-based criteria.

Repeated updates can degrade useful memories~\citep{zhang2026usefulmemoriesfaultycontinuously}, motivating a distinction between initial admission and later maintenance. \methodname{} addresses admission through source-decision evidence.

\runinhead{Decision-aware memory.}
In online content evaluation and social systems, \citet{zhang2026goodrealdiagnosingreducing} identify a gap between AI preference and real user engagement, while social simulation and group-cognition studies---from individual and mean-field simulators~\citep{gao2025s3socialnetworksimulationlarge,mi2025mfllmsimulatingpopulationdecision} to group-agent memory, closed-loop interventions, coupled micro--macro state transitions, and nonlinear group emergence~\citep{zhang2026gas3comprehensivesocialnetwork,zhang2026intervensiminterventionawaresocialnetwork,zhang2026couplingmacrodynamicsmicro,tang2026grouptom}---show that collective trajectories depend on stateful behavioral transitions rather than isolated surface responses. For memory admission, we likewise prioritize behavioral evidence: paired replay measures how a candidate changes the agent's decision distribution.
\citet{liu2021makesgoodincontextexamples} retrieve semantically similar examples to improve in-context learning. EPR uses model likelihoods to supervise demonstration retrieval, addressing context selection rather than persistent writes~\citep{rubin2022learningretrievepromptsincontext}. CICL selects and compresses context using its effects on action choices at retrieval time~\citep{guan2026decisionawarememorycardscounterfactualinspired}. Counterfactual memory auditing of robot policies separates a memory's behavioral influence from whether it guides actions appropriately~\citep{zhang2026memorychangesactionmemory}, and SRP-SFT similarly distinguishes prefix-level distribution matching from selective preference-margin preservation relative to a frozen reference~\citep{wang2026distributionmatchingselectivemargin}. \methodname{} uses paired probes at the source decision and success or failure roles to guide persistent memory admission.

\section{Theoretical Analysis}
\label{app:proofs}

\subsection{Proof of Proposition~\ref{prop:magnitude-bound}}
\label{app:proof-magnitude}

\begin{tcolorbox}[appendixthm]
\textbf{Proposition~\ref{prop:magnitude-bound} (Restated --- Radius bound).}\enspace
\textit{Let $\alpha_t=\operatorname{TV}(P_t,P_t^{m_t})$ and $M_t=D_{\mathrm{JS}}(P_t,P_t^{m_t})$ in base-2 bits. Then $\bigl|\sum_a v_t(a)Q_t(a)\bigr|\leq B_t\alpha_t\leq B_t\sqrt{2\ln2\,M_t}$.}
\end{tcolorbox}

For any constant $c$,
\begin{equation}
\begin{aligned}
&\sum_a v_t(a)Q_t(a)\\
&\quad=\sum_a[P_t^{m_t}(a)-P_t(a)]\,[Q_t(a)-c],
\end{aligned}
\end{equation}
because both distributions have unit mass. Taking $c=(\max_aQ_t(a)+\min_aQ_t(a))/2$ gives
\begin{equation}
\begin{aligned}
\left|\sum_a v_t(a)Q_t(a)\right|
&\leq \frac{B_t}{2}\sum_a|P_t^{m_t}(a)-P_t(a)|\\
&=B_t\operatorname{TV}(P_t,P_t^{m_t})=B_t\alpha_t.
\end{aligned}
\label{eq:range-tv-proof}
\end{equation}
Recall $\bar P_t=(P_t+P_t^{m_t})/2$. Pinsker's inequality in natural logarithms gives
\begin{equation}
\operatorname{TV}(P_t,\bar P_t)^2
\leq\frac12D_{\mathrm{KL}}^{(e)}(P_t\Vert\bar P_t),
\end{equation}
and the same inequality holds for $P_t^{m_t}$. Since
$\operatorname{TV}(P_t,\bar P_t)=\operatorname{TV}(P_t^{m_t},\bar P_t)=\alpha_t/2$, averaging the two inequalities and converting nats to bits yields
\begin{equation}
\alpha_t^2\leq2\ln2\,M_t.
\label{eq:js-tv-proof}
\end{equation}
Combining Eqs.~\ref{eq:range-tv-proof} and~\ref{eq:js-tv-proof} proves Proposition~\ref{prop:magnitude-bound}.\hfill$\square$

\subsection{Probability Transport and Direction}
\label{app:proof-transport}

Recall $v_t(a)=P_t^{m_t}(a)-P_t(a)$ and $\alpha_t=\operatorname{TV}(P_t,P_t^{m_t})$. If $\alpha_t>0$, define
\begin{equation}
\mu_t(a,b)=\frac{[-v_t(a)]_+[v_t(b)]_+}{\alpha_t}.
\label{eq:transport}
\end{equation}
Its row sums are $[-v_t(a)]_+$, its column sums are $[v_t(b)]_+$, and its total mass is $\alpha_t$. Therefore
\begin{equation}
\begin{aligned}
&\sum_{a,b}\mu_t(a,b)[Q_t(b)-Q_t(a)]\\
&\quad=\sum_b[v_t(b)]_+Q_t(b)
 -\sum_a[-v_t(a)]_+Q_t(a)\\
&\quad=\sum_a v_t(a)Q_t(a),
\end{aligned}
\end{equation}
which proves Eq.~\ref{eq:transport-value}. For $\alpha_t=0$, set $\mu_t=0$; both sides are zero.

This identity also establishes direction as independent of magnitude. Consider two memories $m_t^+$ and $m_t^-$ with
\begin{equation}
P_t=(1/2,1/2),\qquad
P_t^{m_t^+}=(3/4,1/4),\qquad
P_t^{m_t^-}=(1/4,3/4),
\end{equation}
and $Q_t=(1,0)$. Symmetry gives
$D_{\mathrm{JS}}(P_t,P_t^{m_t^+})=D_{\mathrm{JS}}(P_t,P_t^{m_t^-})$, while
\begin{equation}
\mathbb E_{P_t^{m_t^\pm}}[Q_t]-\mathbb E_{P_t}[Q_t]=\pm1/4.
\label{eq:equal-js-opposite-value}
\end{equation}
Two memories can therefore have identical shift magnitude and opposite local value.

\subsection{Proof of Proposition~\ref{prop:direction-bound}}
\label{app:proof-direction}

\begin{tcolorbox}[appendixthm]
\textbf{Proposition~\ref{prop:direction-bound} (Restated --- Directional value bound).}\enspace
\textit{Under Assumption~\ref{ass:value-alignment}, $\sum_a v_t(a)Q_t(a)\geq \kappa_t\eta_t - B_t(\alpha_t-\eta_t)$, so memory injection strictly increases expected Q-value whenever $\kappa_t\eta_t>B_t(\alpha_t-\eta_t)$.}
\end{tcolorbox}

Under Assumption~\ref{ass:value-alignment}, if $y_t=+1$, then $v_t(a_t^{(1)})=\eta_t\geq0$, so the transport column ending at $a_t^{(1)}$ has mass exactly $\eta_t$. Every positive-mass transfer in this column gains at least $\kappa_t$ in value. If $y_t=-1$, then $v_t(a_t^{(1)})=-\eta_t\leq0$, so the transport row leaving $a_t^{(1)}$ has mass exactly $\eta_t$, and each of its positive-mass transfers again gains at least $\kappa_t$.

In either case, these transfers contribute at least $\kappa_t\eta_t$. The remaining transport has mass $\alpha_t-\eta_t$, and each unit can lose at most $B_t$. Applying Eq.~\ref{eq:transport-value} gives
\begin{equation}
\sum_a v_t(a)Q_t(a)\geq\kappa_t\eta_t-B_t(\alpha_t-\eta_t).
\end{equation}
When $\kappa_t\eta_t>B_t(\alpha_t-\eta_t)$, the lower bound is strictly positive, proving Proposition~\ref{prop:direction-bound}. If every positive-mass transfer is value-aligned, each summand in Eq.~\ref{eq:transport-value} is nonnegative, and hence $\sum_a v_t(a)Q_t(a)\geq0$.\hfill$\square$

\subsection{From Local Value to Expected Return}
\label{app:local-to-return}

Let $P$ and $P^{\mathrm{mem}}$ be history-dependent policies in the same finite-horizon environment, with bounded rewards, the same initial-history distribution, and terminal value $V_T^P=0$. Memory construction and retrieval may depend on the history; their resulting action distributions are included in the policies. Define
\begin{equation}
\begin{aligned}
Q_t^P(H,a)&=\mathbb E\bigl[r_t+V_{t+1}^P(H_{t+1})\mid H_t=H,a_t=a\bigr],\\
V_t^P(H)&=\sum_a P_t(a\mid H)Q_t^P(H,a).
\end{aligned}
\end{equation}
Conditioning on $H_t=H$ under $P^{\mathrm{mem}}$ yields
\begin{equation}
\begin{aligned}
\mathbb E_{P^{\mathrm{mem}}}\bigl[r_t+V_{t+1}^P(H_{t+1})-V_t^P(H_t)\mid H_t=H\bigr]
&=\sum_a\bigl[P_t^{\mathrm{mem}}(a\mid H)-P_t(a\mid H)\bigr]Q_t^P(H,a)\\
&=\sum_a v_t(H,a)Q_t^P(H,a).
\end{aligned}
\end{equation}
Summing over $t$ and taking expectations cancels the intermediate value terms. Since $\mathbb E[V_0^P(H_0)]=J(P)$, this proves Eq.~\ref{eq:local-to-return}.

At each history reached by $P^{\mathrm{mem}}$, apply the value-alignment condition to $P_t(\cdot\mid H)$ and $P_t^{\mathrm{mem}}(\cdot\mid H)$, using a designated reference action and role and the continuation values $Q_t^P(H,\cdot)$. With the corresponding quantities $\eta_t(H)$, $\alpha_t(H)$, $\kappa_t(H)$, and value range $B_t(H)$, let
\begin{equation}
b_t(H)=\kappa_t(H)\eta_t(H)-B_t(H)[\alpha_t(H)-\eta_t(H)].
\end{equation}
Applying Eq.~\ref{eq:direction-bound} inside the expectation gives
\begin{equation}
J(P^{\mathrm{mem}})-J(P)\geq
\mathbb E_{P^{\mathrm{mem}}}\!\left[\sum_{t=0}^{T-1}b_t(H_t)\right].
\end{equation}
Thus nonnegative $b_t(H)$ at every reachable history implies nonnegative expected-return change. The source-state admission test evaluates local memory effects, while this identity connects effects across retrieved trajectories to task return.\hfill$\square$

\subsection{Information-Theoretic Interpretation}
\label{app:information-view}

For the information-theoretic and finite-sample calculations below, abbreviate $p=P_t$ and $q=P_t^{m_t}$ at a fixed step $t$, so $D_{\mathrm{JS}}(p,q)=M_t$. The magnitude test has an information interpretation. Draw a fair bit $Z\in\{0,1\}$, then draw an action $A$ from $p$ when $Z=0$ and from $q$ when $Z=1$. Writing $m=(p+q)/2$ and using base-2 entropy gives
\begin{equation}
\begin{aligned}
I(Z;A)
&=H(m)-\tfrac12 H(p)-\tfrac12 H(q)\\
&=\tfrac12 D_{\rm KL}(p\Vert m)
 +\tfrac12 D_{\rm KL}(q\Vert m)\\
&=D_{\mathrm{JS}}(p,q).
\end{aligned}
\label{eq:js-mutual-information}
\end{equation}
Thus JS measures how distinguishable the two policy conditions are from their actions; it does not say which actions are better. In particular, $0\leq D_{\mathrm{JS}}(p,q)\leq1$ bit, irrespective of the action-set size.

\paragraph{Coarsening the action space.}
For a deterministic map $g$ from complete actions to action verbs, the Markov chain $Z\to A\to g(A)$ and the chain rule for mutual information yield
\begin{equation}
\begin{aligned}
D_{\mathrm{JS}}(g_{\#}p,g_{\#}q)&=I(Z;g(A))\\
&\leq I(Z;A)=D_{\mathrm{JS}}(p,q).
\end{aligned}
\label{eq:js-coarsening}
\end{equation}
The inequality may be strict: a memory can change which object is selected without changing the verb. This statement concerns the \emph{pushforward of the same complete-action distributions}. The token scorer in Eq.~\ref{eq:appendix-token-score} instead constructs its distribution directly from candidate-prefix likelihoods; Eq.~\ref{eq:js-coarsening} characterizes aggregation of complete-action probabilities.

\paragraph{Local sharpness of the radius bound.}
Take $p=(1/2,1/2)$ and $q=(1/2+\varepsilon,1/2-\varepsilon)$, with $|\varepsilon|<1/2$. The binary entropy expansion $h_2(1/2+x)=1-2x^2/\ln2+O(x^4)$ gives
\begin{equation}
\begin{aligned}
D_{\mathrm{JS}}(p,q)&=\frac{\varepsilon^2}{2\ln2}+O(\varepsilon^4),\\
\operatorname{TV}(p,q)&=|\varepsilon|.
\end{aligned}
\label{eq:js-radius-sharp}
\end{equation}
Consequently, the square-root dependence and leading constant in Eq.~\ref{eq:magnitude-bound} are attained asymptotically for this binary perturbation. The bound controls \emph{capacity} for value change; the transport argument in Appendix~\ref{app:proof-transport} is still needed for its sign.

\subsection{Finite-Sample Resolution of Paired Probes}
\label{app:sample-resolution}

Suppose each of the two fixed policy conditions produces $n$ independent action draws on the same $d\geq2$ categories, with empirical distributions $\widehat p$ and $\widehat q$. This subsection characterizes a sampling-based alternative to token-score probes; it is independent of the token-budget configuration used by the adapters. Hoeffding's inequality and a union bound over the $2d$ empirical coordinates show that, with probability at least $1-\delta$,
\begin{equation}
\max_{x\in\{p,q\}}\max_a|\widehat x(a)-x(a)|
\leq e_n:=\sqrt{\frac{\ln(4d/\delta)}{2n}}.
\label{eq:sample-coordinate}
\end{equation}
Hence each empirical distribution is within $u=\min\{1,de_n/2\}$ in total variation of its population distribution. For the tracked source category $a_t^{(1)}$ and its role $y_t\in\{+1,-1\}$,
\begin{equation}
\bigl|y_t[\widehat q(a_t^{(1)})-\widehat p(a_t^{(1)})]
-y_t[q(a_t^{(1)})-p(a_t^{(1)})]\bigr|\leq2e_n.
\label{eq:sample-margin}
\end{equation}

The classical entropy-continuity inequality~\citep{Audenaert_2007} gives $|H(r)-H(s)|\leq f_d(t)$ when $\operatorname{TV}(r,s)\leq t$, where
\begin{equation}
f_d(t)=
\begin{cases}
h_2(t)+t\log_2(d-1),&t\leq1-1/d,\\
\log_2d,&t>1-1/d.
\end{cases}
\label{eq:entropy-modulus}
\end{equation}
The mixtures $(p+q)/2$ and $(\widehat p+\widehat q)/2$ are also within $u$ in total variation. Applying the entropy bound to the three terms of Eq.~\ref{eq:js-mutual-information} yields
\begin{equation}
|D_{\mathrm{JS}}(\widehat p,\widehat q)-D_{\mathrm{JS}}(p,q)|
\leq 2f_d(u).
\label{eq:sample-js}
\end{equation}
This bound certifies a population JS threshold when the measured value exceeds it by more than $2f_d(u)$; the corresponding signed-margin threshold requires a margin of more than $2e_n$. These certificates cover the scalar magnitude and signed-margin tests, while source-role correctness and top-action switches use their corresponding criteria. Larger sample sizes tighten both certificates.

\section{Experimental Details}
\label{app:experiment-details}

\subsection{Admission Ablation by Task Family}
\label{app:filter-ablation}

Table~\ref{tab:admission-ablation-full} expands the Overall ablation on the same task split. The row at 86.82\% removes failure corrections and retains success-maintain memories; the row at 83.83\% removes success-maintain memories and retains failure corrections. In both rows, the retained role still undergoes its magnitude and directional tests.

\begin{table}[!htbp]
\centering
\appendixfigfont
\anchortablestyle
\setlength{\tabcolsep}{3.5pt}
\renewcommand{\arraystretch}{1.16}
\caption{\textbf{Full ALFWorld admission ablation.} Qwen3.5-27B success (\%) across all six task families. Bold and underline denote best and second-best results.}
\label{tab:admission-ablation-full}
\begin{tabular*}{\textwidth}{@{\extracolsep{\fill}}l*{7}{r}@{}}
\toprule
\tableheader{Admission variant} & \tableheader{Overall}\,$\uparrow$ & \tableheader{Put}\,$\uparrow$ & \tableheader{Clean}\,$\uparrow$ & \tableheader{Heat}\,$\uparrow$ & \tableheader{Cool}\,$\uparrow$ & \tableheader{Examine}\,$\uparrow$ & \tableheader{Puttwo}\,$\uparrow$\\
\midrule
\tablemark{\papericon{anchorInk}{robot}}ReAct (no memory) & 61.19 & 66.67 & 51.61 & 65.22 & 76.19 & 72.22 & 35.29\\
\tablemark{\papericon{anchorInk}{layer-group}}\ablnobehavior{} & 73.38 & 77.78 & 70.97 & 68.12 & 82.54 & 85.19 & 54.90\\
\addlinespace[2pt]
\tablemark{\amplitudermark}\ablnoaspecter{} & 74.88 & 81.94 & 69.89 & 73.91 & 80.95 & 83.33 & 58.82\\
\tablemark{\successmark}\ablnofailure{} & \second{86.82} & \second{91.67} & \second{86.02} & \second{79.71} & 90.48 & \second{92.59} & \second{80.39}\\
\tablemark{\failuremark}\ablnosuccess{} & 83.83 & 88.89 & 82.80 & 76.81 & \second{92.06} & 88.89 & 72.55\\
\midrule
\rowcolor{oursgray}
\anchorrowname & \best{96.27} & \best{100.00} & \best{90.32} & \best{91.30} & \best{100.00} & \best{100.00} & \best{100.00}\\
\bottomrule
\end{tabular*}
\end{table}

\subsection{BankMaster Details}
\label{app:bankmaster-details}

BankMaster stores each admitted $m_t$ verbatim in the card's \texttt{memory\_guidance} field and adds role, applicability conditions, retrieval keys, and source metadata. A success hint specifies an action to reinforce under matching conditions; a failure hint records a pattern to avoid and its immediate correction. Card construction enriches metadata without rewriting, summarizing, or otherwise changing the tested guidance.

The retrieval score in Eq.~\ref{eq:bankmaster-retrieval} combines task-family and state similarity with action compatibility. Up to three cards above threshold $\tau_R$ enter the policy context; cards whose cues or recommended actions do not match the decision are ignored. Table~\ref{tab:implementation-parameters} lists the retrieval threshold, cross-family settings, skip conditions, and card-context budget.

Each card stores confidence, support count, source metadata, and reuse statistics. After an episode, BankMaster updates retrieved-card confidence from feedback, merges similar cards with the same role, decays stale records, and prunes low-confidence cards at capacity, as formalized in Eq.~\ref{eq:bankmaster-update}. During held-out evaluation, the constructed bank supplies cards without updates.

\subsection{Implementation Parameters}
\label{app:implementation-parameters}

\begin{table}[!htbp]
\centering
\appendixfigfont
\anchortablestyle
\setlength{\tabcolsep}{4.5pt}
\renewcommand{\arraystretch}{1.16}
\caption{\textbf{Implementation settings.} Branch-specific thresholds use JS in bits. The measurement interface is described in Appendix~\ref{app:prototype-delta}.}
\label{tab:implementation-parameters}
\begin{tabular*}{\textwidth}{@{\extracolsep{\fill}}>{\raggedright\arraybackslash}p{.205\textwidth}>{\raggedright\arraybackslash}p{.255\textwidth}>{\raggedright\arraybackslash}p{.205\textwidth}>{\raggedright\arraybackslash}p{.255\textwidth}@{}}
\toprule
\tableheader{Setting} & \tableheader{Value} & \tableheader{Setting} & \tableheader{Value}\\
\midrule
\rowcolor{groupgray}
\multicolumn{4}{@{}l@{}}{\tablegroup{crosshairs}{Decoding \& Paired Replay Probe}}\\
Action decoding & Token-wise greedy (256 tok.) & Probe statistic & Token-level action scores\\
Replay token budget $L$ & ALFWorld: 1; WebShop: 5 & Paired context & Fixed within each pair\\
Localized candidates & $K=5$ per episode & Maximum cards & 8 per episode\\
\addlinespace[2pt]
\rowcolor{groupgray}
\multicolumn{4}{@{}l@{}}{\tablegroup{compass}{Magnitude \& Source-Role Admission Gate}}\\
Failure JS threshold & $\tau_{\mathrm{fail}}=0.12$ bits & Success JS threshold & $\tau_{\mathrm{succ}}=0.05$ bits\\
Failure probability branch & Source decrease $\geq0.10$ & Failure ranking branch (OR) & Injected top action $\neq$ source\\
Success ranking & Injected top action $=$ source & Success reinforcement & Increase $\geq0.04$\\
\addlinespace[2pt]
\rowcolor{groupgray}
\multicolumn{4}{@{}l@{}}{\tablegroup{database}{BankMaster Retrieval \& Maintenance}}\\
Retrieval size & Top 3 cards & Retrieval threshold & $\tau_R=0.45$\\
Cross-family retrieval & Boost $0.50$, penalty $0.15$ & Card context budget & 320 tokens per card\\
Retrieval skip rule & Conf.\ $\geq0.50$, ent.\ $\leq1.40$ or margin $\geq0.12$ & Bank capacity / merge & 2,000 cards / $0.85$ sim.\\
\bottomrule
\end{tabular*}
\end{table}

\runinhead{ALFWorld data.}
We use the raw TextWorld/PDDL game files in \href{https://huggingface.co/datasets/awawa-agi/alfworld-raw/tree/main/data}{\texttt{awawa-agi/alfworld-raw}}, including training and evaluation game specifications. Task counts below refer to the selected experimental tasks.

\runinhead{ALFWorld evaluation set.}
The held-out ALFWorld set contains 402 tasks: 72 Put, 93 Clean, 69 Heat, 63 Cool, 54 Examine, and 51 Puttwo. Overall success pools all evaluated tasks. For task family $f$, let $s_f$ and $n_f$ denote its success count and evaluation count. We report
\begin{equation}
\mathrm{Overall}=100\,\frac{\sum_f s_f}{\sum_f n_f},\qquad
\mathrm{Success}_f=100\,\frac{s_f}{n_f}.
\end{equation}
Thus Overall is the total number of successes divided by the total evaluation count, rather than an unweighted average of task-family percentages.

\runinhead{WebShop task split.}
We use the 10,000-product subset of \href{https://huggingface.co/datasets/YWZBrandon/webshop-data/tree/main/subsets/10k}{\texttt{YWZBrandon/webshop-data}}.
Product-ID matching yields 4,455 attributed instructions (400 construction; 4,055 evaluation); 5,545 products have none. Counts refer to product-linked instructions, not unique strings or all product-option combinations.

\runinhead{WebShop evaluation metric.}
WebShop is evaluated by mean score, not success rate. Let $s_i^{\rm WS}$ denote the score of evaluation task $i$ on the reported score scale. Over $N_{\rm WS}=4{,}055$ evaluation tasks, we compute
\begin{equation}
\mathrm{MeanScore}_{\rm WS}=\frac{1}{N_{\rm WS}}\sum_{i=1}^{N_{\rm WS}}s_i^{\rm WS}.
\end{equation}
Unlike ALFWorld's binary success indicator, this metric averages the task scores directly. Mean interaction steps are reported separately as an efficiency metric.

\subsection{Admission Configuration and Measurement}
\label{app:prototype-delta}

\paragraph{Measurement interface.}
The two adapters share a token-budget parameter $L$, counting action-content tokens immediately after \texttt{<action>}. The opening tag is context, not part of the scored span. ALFWorld scores the action head ($L=1$); WebShop scores up to five action tokens ($L=5$). Appendix~\ref{app:prompts} provides one replay template per environment: the complete saved decision context, with its history, task guidance, and retrieved memory. Only the injected condition adds $m_t$. The scoring specification below applies to the paired measurements, including the case studies in Appendix~\ref{app:cases}.

\paragraph{Candidate actions.}
At each source decision, store a finite ordered list $\mathcal C_t$ of legal candidate action strings, including the recorded source action. ALFWorld obtains this list from the TextWorld environment's \texttt{admissible\_commands}. WebShop obtains clickable products, option values, and navigation controls from the current page's \texttt{get\_available\_actions()} interface, and serializes each as a complete \texttt{click[...]} command. Thus product selection and option selection are distinct candidates, not a single \texttt{click} category. The environment interfaces are documented in the \href{https://github.com/alfworld/alfworld}{ALFWorld} and \href{https://github.com/princeton-nlp/WebShop}{WebShop} releases. When search is available, the frozen policy generates a finite list of query candidates from the current task and page before the paired test. The adapter serializes these as \texttt{search[query]} actions and stores the list with the source snapshot. Replay reuses this list and never generates new candidates from the injected hint or subsequent feedback. Remove duplicate complete strings while preserving their first occurrence.

\paragraph{Token scores and shared prefixes.}
To score candidates, tokenize each action continuation using the policy tokenizer and the exact saved prefix ending in \texttt{<action>}. Let $z(a)=(z_1,\ldots,z_{n(a)})$ be its action-content tokens, excluding the opening and closing tags, and let $k(a)=\min\{L,n(a)\}$. Define $\phi_L(a)=z_{1:k(a)}$ and form the set of distinct scored sequences
\begin{equation}
\mathcal V_t=\{\phi_L(a):a\in\mathcal C_t\}.
\end{equation}
Candidates with identical scored sequences share one category. This deduplication prevents a verb from gaining weight merely because it has more object-level completions. For each $v=(v_1,\ldots,v_{|v|})\in\mathcal V_t$, teacher-force its tokens under each saved context $h\in\{h_t,h_t^{m_t}\}$ and sum their log probabilities:
\begin{equation}
\begin{aligned}
\ell_L(v\mid h)&=\sum_{j=1}^{|v|}\log\pi_\theta\bigl(v_j\mid h,\texttt{<action>},v_{<j}\bigr),\\
\widetilde P_h(v)&=\frac{\exp\ell_L(v\mid h)}{\sum_{u\in\mathcal V_t}\exp\ell_L(u\mid h)}.
\end{aligned}
\label{eq:appendix-token-score}
\end{equation}
We sum token log probabilities, using all content tokens for actions shorter than $L$. Leading whitespace and token IDs follow the same saved serialization in both conditions. The resulting distribution is normalized over distinct candidate prefixes: actions sharing their first $L$ tokens form one category. With $L=5$, WebShop click targets occupy different categories when their tokenized arguments differ within the scored span.

The source category is $v_t^{(1)}=\phi_L(a_t^{(1)})$. Compute JS and the source-category change using $\widetilde P_{h_t}$ and $\widetilde P_{h_t^{m_t}}$ on exactly $\mathcal V_t$. For example, if \texttt{take apple 1} and \texttt{take mug 2} share the same first token, $L=1$ assigns both to one category; that category appears only once in the softmax. A valid pair requires the saved context, source candidate, identical tokenization, and finite log probabilities for every scored token in both conditions. Missing top-$k$ scores invalidate the pair rather than being treated as zero. Resolve score ties by the stored candidate order.

\paragraph{Admission configuration.}
Both source roles pass a magnitude test and a directional test. We use $\tau_{\rm succ}=0.05$ bits for success-maintain memories and $\tau_{\rm fail}=0.12$ bits for failure corrections. Success-maintain memories require the injected top category to equal the source category and its probability to increase by at least $\epsilon_{\rm succ}=0.04$. Failure corrections require either a source-category decrease of at least $0.10$ or an injected top category different from the failed source category. Algorithm~\ref{alg:branch-admission} gives the complete predicates.

\paragraph{Baseline adaptation.}
Each comparison uses the backbone and evaluation split specified in Table~\ref{tab:main-results}. For a baseline whose open-source release includes the corresponding benchmark environment, we use that implementation. Otherwise, we adapt its memory mechanism to the same environment and action-selection prompt as \methodname{}, including the task guidance supplied to the policy. Each baseline retains its own memory extraction, organization, and retrieval procedure. These are complete-system comparisons; Table~\ref{tab:admission-ablation} isolates admission variants within the \methodname{} pipeline.

\paragraph{Token accounting.}
The WebShop panel in Figure~\ref{fig:memory-efficiency} uses Qwen3.5-27B and counts input and output tokens across all model calls during both memory construction and held-out evaluation. For each system, let $\mathcal Q$ be this complete call set. Total consumption is
\begin{equation}
T(\mathcal Q)=\sum_{q\in\mathcal Q}\bigl(n_{\mathrm{in}}(q)+n_{\mathrm{out}}(q)\bigr).
\end{equation}
The total includes action selection, candidate-query generation, reflection and localization, hint writing, both replay conditions, card construction, and model-based maintenance wherever these calls occur in the system. Repeated history and retrieved-card tokens count on each call in which they occur; $L$ is the scoring span, not the total input length. Environment steps and local, non-model retrieval are not model tokens. Baselines are counted over the same construction-plus-evaluation scope, including their own memory-related model calls. The per-task labels divide this total by 4,055 evaluation tasks, amortizing construction cost across evaluation. Circle area is proportional to the mean evaluation steps reported in Table~\ref{tab:main-results}.

\paragraph{Source roles.}
The Localizer receives the goal, the recorded trajectory, and episode feedback, and autonomously assigns each selected source step a hindsight credit $c_t\in[-1,1]$. Positive credit identifies a decision to reinforce, negative credit identifies a decision to correct, and zero credit denotes an uncertain role. For nonzero credit, the step role is $y_t=\operatorname{sign}(c_t)$. A success-maintain memory identifies a local decision to reinforce; a failure correction identifies a local decision to avoid or correct. The role describes the step, not the outcome of the whole episode. The categorical mapping is $y_t=+1$ for \texttt{success\_maintain} and $y_t=-1$ for \texttt{failure\_correction}; \texttt{uncertain} candidates are excluded from admission. The gate uses this sign, not the magnitude of $c_t$. The prompt's \texttt{importance\_score} ranks candidates and is separate from signed hindsight credit. Task-specific guidance supplied to the Localizer is reproduced in Appendix~\ref{app:prompts}.

\paragraph{Evaluation signals.}
Paired replay measures a fixed hint's effect on the normalized candidate-prefix distribution. Local verification refers to passing the magnitude and role-specific preference predicates on this distribution. The theoretical bounds instead concern complete-action distributions and condition on Q-value gaps and residual mass transfer. These conditions are analytical assumptions, not quantities measured by the gate. In particular, the failure ranking branch can admit an alternative top category even when the source probability increases; it tests a different preference criterion from positive signed-margin alignment. Held-out task results assess the constructed bank's utility under the deployed decoding rule.

\subsection{Mechanistic Metrics}
\label{app:mechanistic-metrics}
We log candidate roles, JS, signed margins $\eta_t$, action changes, and rejection reasons, along with episode-level candidate counts, memory costs, probes, and outcomes. Figure~\ref{fig:memory-efficiency} reports cumulative success $S(N)=100N^{-1}\sum_{i=1}^{N}\mathbf 1\{\text{episode }i\text{ succeeds}\}$ over 200 randomly sampled ALFWorld construction tasks, for $1\leq N\leq200$; dashed curves show bank size. Each point summarizes the task stream up to that episode, with the bank evolving during construction. It is a construction-stream statistic rather than repeated evaluation of bank checkpoints on a fixed task set. Table~\ref{tab:main-results} evaluates the final banks on the separate held-out set.

\section{Algorithms}
\label{app:algorithm}

Algorithm~\ref{alg:anchor-loop} makes the write-and-reuse loop explicit. During memory construction, the agent extracts candidates after observing an episode's feedback. During held-out evaluation, the constructed bank is read without admitting memories from evaluation outcomes. Local replay scores a saved decision; it does not execute an alternative trajectory.

\par\smallskip\noindent
\begin{minipage}{\textwidth}
\captionsetup{type=anchoralgorithm}
\begin{tcolorbox}[appendixalgorithm,title={\anchormark\hspace{.45em}Algorithm~\ref{alg:anchor-loop}: Memory Evolution Loop},
  top=4pt,bottom=4pt,before skip=2pt,after skip=2pt]
\appendixfigfont
\textbf{Input:} frozen policy $\pi_\theta$; task stream; initial bank $\mathcal B$; admission rule $G$; candidate budget $K$; write flag $w$.\par\vspace{1.5pt}\hrule\vspace{2.5pt}
\begin{enumerate}[label={\color{anchorInk!70}\scriptsize\textbf{\arabic*:}},leftmargin=1.8em,itemsep=.6pt,parsep=0pt,topsep=1pt]
\item \textbf{for each} task $g$ \textbf{do}
\item \quad Initialize history $H$ and trajectory $\tau$.
\item \quad \textbf{while} the episode is active \textbf{do}
\item \qquad Observe $o$ and admissible $\mathcal A$; retrieve $M\gets\mathcal R(\mathcal B,H)$.
\item \qquad Build and save $h=F(H,\mathcal A,M)$; select and execute $a$ from $\pi_\theta$.
\item \qquad Update $H$; append the transition and feedback to $\tau$.
\item \quad \textbf{end while}; record outcome $z$.
\item \quad \textbf{if} $w=0$ \textbf{then continue}.
\item \quad $\mathcal M(\tau)\gets\localizer(\tau)$; keep at most $K$ candidates.
\item \quad \textbf{for each} $m_t\in\mathcal M(\tau)$ \textbf{do}
\item \qquad Load $h_t,\mathcal A_t,a_t^{(1)},y_t$; freeze $m_t$.
\item \qquad Add only $m_t$ to a copy of $h_t$; probe both on fixed candidates.
\item \qquad \textbf{if} probes are invalid: \textbf{continue}.
\item \qquad Compute $M_t,\Delta p_t^{(1)}$ (Eqs.~\ref{eq:observation-js} and~\ref{eq:observation-direction}).
\item \qquad \textbf{if} $G_t=1$ (Eq.~\ref{eq:method-admission}): finalize a card and insert it into $\mathcal B$.
\item \quad \textbf{end for}; update confidence from feedback; decay, merge, and prune $\mathcal B$.
\item \textbf{end for}; \textbf{return} outcomes and $\mathcal B$.
\end{enumerate}
\end{tcolorbox}
\captionof{anchoralgorithm}{\textbf{Memory evolution.} Construction episodes ($w=1$) extract, locally verify, and maintain cards; held-out evaluation ($w=0$) performs read-only retrieval.}
\label{alg:anchor-loop}
\end{minipage}
\par\medskip

\subsection{Formal Admission}
\label{app:formal-admission}
For the action distributions in Section~\ref{sec:method}, $G_t$ is the branch-specific admission test in Eq.~\ref{eq:method-admission}:
\begin{equation}
\begin{aligned}
y_t=+1\text{ (success maintain):}\quad & M_t\geq\tau_{\rm succ}\ \land\ \widehat a_t^{m_t}=a_t^{(1)}\ \land\ \Delta p_t^{(1)}\geq\epsilon_{\rm succ},\\
y_t=-1\text{ (failure correction):}\quad & M_t\geq\tau_{\rm fail}\ \land\ \bigl(\Delta p_t^{(1)}\leq-\epsilon_{\rm drop}\ \lor\ \widehat a_t^{m_t}\neq a_t^{(1)}\bigr).
\end{aligned}
\end{equation}
The action set, source history, and previously retrieved memory are matched across probes; only the candidate hint changes.

\subsection{Branch-Specific Implementation}
\label{app:algorithm-predicates}
The token-budgeted gate applies Eq.~\ref{eq:method-admission} to $\widetilde P_t=\widetilde P_{h_t}$ and $\widetilde P_t^{m_t}=\widetilde P_{h_t^{m_t}}$ on the fixed prefix categories $\mathcal V_t$:
\begin{equation}
M_t=D_{\rm JS}(\widetilde P_t,\widetilde P_t^{m_t}),\qquad
\Delta p_t^{(1)}=\widetilde P_t^{m_t}(v_t^{(1)})-\widetilde P_t(v_t^{(1)}),\qquad
b_t^{m_t}=\arg\max_{v\in\mathcal V_t}\widetilde P_t^{m_t}(v).
\end{equation}

\par\smallskip\noindent
\begin{minipage}{\textwidth}
\captionsetup{type=anchoralgorithm}
\begin{tcolorbox}[appendixalgorithm,title={\aspectermark\hspace{.45em}Algorithm~\ref{alg:branch-admission}: Branch-Specific Admission Gate},
  top=4pt,bottom=4pt,before skip=2pt,after skip=2pt]
\appendixfigfont
\textbf{Input:} paired distributions $(\widetilde P_t,\widetilde P_t^{m_t})$; shift $M_t$; source category $v_t^{(1)}$; probability change $\Delta p_t^{(1)}$; role $y_t$.\par\vspace{1.5pt}\hrule\vspace{2.5pt}
\begin{enumerate}[label={\color{anchorInk!70}\scriptsize\textbf{\arabic*:}},leftmargin=1.8em,itemsep=1pt,parsep=0pt,topsep=1.5pt]
\item \textbf{if} either probe is invalid: \textbf{return false}.
\item \textbf{if} the role is failure correction ($y_t=-1$):
\item \quad \textbf{return} $M_t\geq\tau_{\rm fail}$ \textbf{and} $(\Delta p_t^{(1)}\leq-\epsilon_{\rm drop}$ \textbf{or} $b_t^{m_t}\neq v_t^{(1)})$.
\item \textbf{if} the role is success-maintain memory ($y_t=+1$):
\item \quad \textbf{return} $M_t\geq\tau_{\rm succ}$ \textbf{and} $b_t^{m_t}=v_t^{(1)}$ \textbf{and} $\Delta p_t^{(1)}\geq\epsilon_{\rm succ}$.
\item \textbf{return false} for an uncertain role ($y_t=0$).
\end{enumerate}
\end{tcolorbox}
\captionof{anchoralgorithm}{\textbf{Branch-specific gate.} Combines Amplituder JS thresholds ($\tau_{\rm succ},\tau_{\rm fail}$) with Aspecter role-aligned preference checks.}
\label{alg:branch-admission}
\end{minipage}
\par\medskip
The thresholds are $\tau_{\rm succ}=0.05$, $\tau_{\rm fail}=0.12$, $\epsilon_{\rm drop}=0.10$, and $\epsilon_{\rm succ}=0.04$. The lower success threshold allows smaller shifts when the baseline already supports the source action. The failure ranking branch compares the injected top category with the fixed source category, independently of the baseline top category. The probability branch measures suppression; the ranking branch checks which category is preferred after injection.

\par\smallskip\noindent
\begin{tcolorbox}[appendixrecord,title={\amplitudermark\hspace{.45em}Summary of Gate Thresholds and Design Rationale}]
\renewcommand{\arraystretch}{1.15}
\begin{tabular*}{\linewidth}{@{\extracolsep{\fill}}lll>{\raggedright\arraybackslash}p{.48\linewidth}@{}}
\toprule
\tableheader{Symbol} & \tableheader{Role branch} & \tableheader{Value} & \tableheader{Behavioral interpretation}\\
\midrule
$\tau_{\mathrm{succ}}$ & Success maintain ($y_t=+1$) & $0.05$ bits & Admits modest shifts when baseline already favors $v_t^{(1)}$\\
$\epsilon_{\mathrm{succ}}$ & Success maintain ($y_t=+1$) & $+0.04$ & Requires positive reinforcement while keeping $b_t^{m_t}=v_t^{(1)}$\\
$\tau_{\mathrm{fail}}$ & Failure correction ($y_t=-1$) & $0.12$ bits & Requires decisive distribution movement away from error\\
$\epsilon_{\mathrm{drop}}$ & Failure correction ($y_t=-1$) & $-0.10$ & Accepts source suppression or top-category switch ($b_t^{m_t}\neq v_t^{(1)}$)\\
\bottomrule
\end{tabular*}
\end{tcolorbox}

\section{Prompt Templates}
\label{app:prompts}

Figure~\ref{fig:prompt-templates} collects the complete prompts for five stages: action selection, memory extraction, hint construction, paired replay, and card construction. Braced fields are filled at runtime. The templates retain all environment-specific instructions, including the ALFWorld heating protocol.

\newcommand{\promptfield}[1]{\par\vspace{1.5pt}\noindent{\bfseries\color{anchorInk} #1.}\enspace}

\paragraph{Information flow.}
The Localizer uses the trajectory and episode feedback to assign hindsight roles. The hint writer describes a local action under the source decision's conditions, without presenting future observations as already available or supplying a complete solution. Candidate text remains fixed during replay. After admission, the card builder copies it verbatim into \texttt{memory\_guidance} and adds only retrieval and source metadata.

\paragraph{Paired context.}
Before executing an action, save the rendered messages: system instructions, history selected by $F$, task guidance, observation, ordered admissible actions, and retrieved cards. Replay clones this snapshot twice and adds \textbf{CANDIDATE GUIDANCE} immediately after the memory-evidence block in the injected copy. Both calls retain the original card text and order, history clipping, model, tokenizer, candidate actions, output contract, and $L$; inserting the hint must not truncate another field. The baseline omits only the candidate paragraph. Existing memory evidence is absent from both calls only when absent from the source prompt. The source action and hindsight role remain checker metadata; subsequent feedback is supplied only to the Localizer and hint writer. A missing snapshot or required score invalidates the pair.

\par\smallskip\noindent
\begin{minipage}[t]{.485\textwidth}
\vspace{0pt}
\subsection{ALFWorld Prompts}
\label{app:alfworld-prompts}
\begin{tcolorbox}[appendixrecord,equal height group=prompt-env-summary,
  title={\alfmark\hspace{.4em}ALFWorld Specification (Panels a--e)}]
\textbf{Action \& replay (a, d).} Same action-only output contract with saved goal, family recipe, phase guidance, observation, admissible actions, history, and cards. Both request a complete admissible action; replay scores the first content token after \texttt{<action>} ($L=1$).\par\smallskip
\textbf{Extraction \& storage (b, c, e).} Localizer selects $\leq 5$ consequential steps; long rollouts compact to $\leq 20$ steps with clipped observations. Panel~(c) redirects away from failures; success hints reinforce the observed action without future steps. Card builder copies admitted hints verbatim into \texttt{memory\_guidance}~(e).
\end{tcolorbox}
\end{minipage}\hfill
\begin{minipage}[t]{.485\textwidth}
\vspace{0pt}
\subsection{WebShop Prompts}
\label{app:webshop-prompts}
\begin{tcolorbox}[appendixrecord,equal height group=prompt-env-summary,
  title={\webmark\hspace{.4em}WebShop Specification (Panels f--j)}]
\textbf{Action \& replay (f, i).} Supplies instruction, task family, page phase, history (clipped to 1,800 chars), current page (5,000 chars), available actions, and cards. Replay reuses the exact rendered fields and scores up to $L=5$ action-content tokens on fixed candidates.\par\smallskip
\textbf{Extraction \& storage (g, h, j).} Localizer selects $\leq 5$ shopping decisions (product, option, checkout); observations clip to 400/300 chars. Hint writer gives a local check or reinforcement~(h); card builder preserves the verified hint verbatim and adds phase, attribute, and retrieval metadata~(j).
\end{tcolorbox}
\end{minipage}
\par\medskip

\begin{tcolorbox}[appendixbox,title={\anchormark\hspace{.45em}Five-Stage Prompt Atlas Index (Figure~\ref{fig:prompt-templates})}]
\appendixfigfont
\anchortablestyle
\renewcommand{\arraystretch}{1.15}
\begin{tabular*}{\linewidth}{@{\extracolsep{\fill}}lll>{\raggedright\arraybackslash}p{.46\linewidth}@{}}
\toprule
\tableheader{Stage} & \tableheader{ALFWorld} & \tableheader{WebShop} & \tableheader{Runtime contract \& invariant}\\
\midrule
1. Action selection & Panel (a) & Panel (f) & Emits one verbatim action inside \texttt{<action>...</action>}\\
2. Memory extraction & Panel (b) & Panel (g) & Ranks $\leq 5$ steps with signed hindsight credit $c_t\in[-1,1]$\\
3. Hint construction & Panel (c) & Panel (h) & Writes a local check-only hint without future leakage\\
4. Paired replay & Panel (d) & Panel (i) & Clones source context; injects \texttt{CANDIDATE GUIDANCE} ($L=1$ / $5$)\\
5. Card construction & Panel (e) & Panel (j) & Copies verified hint verbatim into \texttt{memory\_guidance}\\
\bottomrule
\end{tabular*}
\end{tcolorbox}

% Seamless non-floating prompt atlas: flows continuously across pages without large bottom gaps.
\par\medskip\noindent
\begin{minipage}{\textwidth}
\captionsetup{type=figure}
\begin{tcolorbox}[appendixprompt,title={\alfmark\hspace{.4em}(a) ALFWorld / 1. Action selection\hfill\textit{Role: User}}]
\promptkey{TASK:} \promptvar{goal}\quad $\cdot$\quad \promptvar{family\_recipe}\quad $\cdot$\quad \promptvar{phase\_hint} \promptvar{task\_specific\_protocol}
\promptfield{Memory-use rules}Treat each retrieved card as one local state transition, never as a complete plan. Match its task family, exact target noun, inventory cue, phase, and current admissible action. A matching success card may support the listed action. A fail card only describes a pattern to avoid and a local correction; never copy its failed action. Ignore a card when any of these cues conflict.
\par\smallskip
\promptkey{MEMORY DECISION EVIDENCE:} \promptvar{retrieved\_cards} Use it only when its cue and exact available action match the current state.
\promptfield{Action rules}Use only exact actions from Admissible actions. Do not invent new object names or container names. Do not choose \texttt{look}, \texttt{inventory}, or \texttt{help} unless no task-progress action is available.
\par\smallskip
\promptkey{RECENT ACTION HISTORY:} \promptvar{recent\_history}\par\vspace{1pt}
\promptkey{Latest observation:} \promptvar{observation}\par\vspace{1pt}
\promptkey{Admissible actions:} \promptvar{admissible\_actions}
\promptfield{Output}Return exactly one complete action inside \texttt{<action>...</action>}. Copy the action verbatim from Admissible actions. Output no explanation.
\end{tcolorbox}
\captionof{figure}{\textbf{Complete prompt templates (Panels a--j across ALFWorld and WebShop).} Braced fields are filled at runtime; Panels~(a)--(e) specify ALFWorld and Panels~(f)--(j) specify WebShop.}
\label{fig:prompt-templates}
\end{minipage}
\par\smallskip

\begin{tcolorbox}[appendixprompt,title={\alfmark\hspace{.4em}(b) ALFWorld / 2. Memory extraction\hfill\textit{Role: User}}]
\promptfield{Role and task}You are \texttt{CriticalStepLocator}. Return JSON only. Select at most 5 most important decision steps, ranked by \texttt{importance\_score}. Avoid routine navigation and duplicate observations.
\promptfield{Output schema}Each item must contain \texttt{step\_id}, \texttt{c\_t}, \texttt{critical\_role}, \texttt{importance\_score}, \texttt{importance\_type}, \texttt{why\_critical}, \texttt{expected\_fix\_direction}. Set \texttt{critical\_role} to exactly one of \texttt{success\_maintain}, \texttt{failure\_correction}, or \texttt{uncertain}. Use the top-level JSON key \texttt{candidates}.
\promptfield{Hindsight credit}Independently score each selected source decision with \texttt{c\_t} in $[-1, 1]$, using the goal, trajectory, and episode feedback. Assign a positive score to a local action that should be reinforced and a negative score to an action that should be corrected. Set \texttt{c\_t} to 0 and \texttt{critical\_role} to \texttt{uncertain} when the evidence does not support either role. For nonzero \texttt{c\_t}, its sign must agree with \texttt{critical\_role}: positive means \texttt{success\_maintain} and negative means \texttt{failure\_correction}. Judge the local step rather than copying the whole-episode outcome. Keep \texttt{importance\_score} as the separate candidate-ranking score.
\promptfield{Selection rules}For a failed episode, prioritize the earliest wrong local decision that could have been corrected, and include at least one \texttt{failure\_correction} when the rollout contains an obvious detour, \texttt{help}/\texttt{look} misuse, unrelated object interaction, or skipped task-progress action.
\promptfield{Environment rule: heating}For a heat task, treat the microwave protocol as distinct steps: take the target, go to microwave, heat it, then go to the destination and move it. The microwave itself is not opened or closed for this task.
\par\smallskip
\promptkey{EPISODE OUTCOME:} \promptvar{episode\_outcome}\quad $\cdot$\quad \promptkey{GOAL:} \promptvar{goal}\par\vspace{1pt}
\promptkey{ROLLOUT:} \promptvar{trajectory\_json}
\end{tcolorbox}

\begin{tcolorbox}[appendixprompt,title={\alfmark\hspace{.4em}(c) ALFWorld / 3. Hint construction\hfill\textit{Role: User}}]
\promptfield{Role and task}You are \texttt{ANCHOR write\_hint}. Return JSON only with \texttt{hint\_text}, \texttt{hint\_type}, and \texttt{leakage\_risk}.
\promptfield{Guidance rules}Write a local check-only action hint. Do not reveal future observations or a full solution. This is a \texttt{failure\_correction} candidate. The hint must explicitly redirect away from the observed action toward the immediate local correction in \texttt{expected\_fix\_direction}; do not describe the observed action as correct.
\par\smallskip
\promptkey{GOAL:} \promptvar{goal}\quad $\cdot$\quad \promptkey{CANDIDATE:} \promptvar{candidate\_json}\quad $\cdot$\quad \promptkey{STATE:} \promptvar{state\_json}
\end{tcolorbox}

\begin{tcolorbox}[appendixprompt,title={\alfmark\hspace{.4em}(d) ALFWorld / 4. Paired replay\hfill\textit{Role: User}}]
\promptkey{TASK:} \promptvar{goal}\quad $\cdot$\quad \promptvar{family\_recipe}\quad $\cdot$\quad \promptvar{phase\_hint} \promptvar{task\_specific\_protocol}
\promptfield{Memory-use rules}Treat each retrieved card as one local state transition, never as a complete plan. Match its task family, exact target noun, inventory cue, phase, and current admissible action. A matching success card may support the listed action. A fail card only describes a pattern to avoid and a local correction; never copy its failed action. Ignore a card when any of these cues conflict.
\par\smallskip
\promptkey{MEMORY DECISION EVIDENCE:} \promptvar{retrieved\_cards} Use it only when its cue and exact available action match the current state.\par\vspace{1pt}
\promptkey{CANDIDATE GUIDANCE:} \promptvar{candidate\_hint} Apply this guidance only when it matches the current state and an admissible action.
\promptfield{Action rules}Use only exact actions from Admissible actions. Do not invent new object names or container names. Do not choose \texttt{look}, \texttt{inventory}, or \texttt{help} unless no task-progress action is available.
\par\smallskip
\promptkey{RECENT ACTION HISTORY:} \promptvar{recent\_history}\par\vspace{1pt}
\promptkey{Latest observation:} \promptvar{observation}\par\vspace{1pt}
\promptkey{Admissible actions:} \promptvar{admissible\_actions}
\promptfield{Output}Return exactly one complete action inside \texttt{<action>...</action>}. Copy the action verbatim from Admissible actions. Output no explanation.
\end{tcolorbox}

\begin{tcolorbox}[appendixprompt,title={\alfmark\hspace{.4em}(e) ALFWorld / 5. Card construction\hfill\textit{Role: User}}]
\promptfield{Role and task}You are \texttt{ANCHOR MemoryCardBuilder}. Return JSON only. Build a reusable step-level card without future-only information or a full solution.
\promptfield{Role mapping}The candidate \texttt{critical\_role} defines the card type: \texttt{success\_maintain} means \texttt{success\_memory}, while \texttt{failure\_correction} means \texttt{fail\_memory}. This step type is independent of whether the whole episode succeeded; a successful episode can contain a \texttt{fail\_memory} and a failed episode can contain a \texttt{success\_memory}.
\promptfield{Card fields}Include \texttt{goal\_signature}, \texttt{visible\_entities}, \texttt{inventory\_state}, \texttt{memory\_guidance}, \texttt{task\_plan}, \texttt{retrieval\_keys}, \texttt{confidence}, \texttt{support\_count}, \texttt{uses\_future\_information}, \texttt{contains\_full\_solution}, \texttt{copies\_verification\_hint}.
\promptfield{Guidance preservation}Set \texttt{memory\_guidance} to \texttt{VERIFIED HINT} verbatim and \texttt{copies\_verification\_hint} to \texttt{true}. Add only retrieval and source metadata; do not paraphrase, summarize, or extend the guidance.
\par\smallskip
\promptkey{VERIFIED HINT:} \promptvar{candidate\_hint}\quad $\cdot$\quad \promptkey{FAMILY:} \promptvar{task\_family}\quad $\cdot$\quad \promptkey{GOAL:} \promptvar{goal}\par\vspace{1pt}
\promptkey{CANDIDATE:} \promptvar{candidate\_json}\quad $\cdot$\quad \promptkey{STATE:} \promptvar{state\_json}\quad $\cdot$\quad \promptkey{POLICY\_DIFF:} \promptvar{policy\_diff\_json}\quad $\cdot$\quad \promptkey{TASK\_PLAN:} \promptvar{task\_plan\_json}
\end{tcolorbox}

\begin{tcolorbox}[appendixpromptws,title={\webmark\hspace{.4em}(f) WebShop / 1. Action selection\hfill\textit{Role: User}}]
\promptkey{TASK:} \promptvar{goal}\quad $\cdot$\quad \promptkey{TASK FAMILY:} \promptvar{task\_family}\quad $\cdot$\quad \promptkey{PHASE GUIDANCE:} \promptvar{current\_phase}
\promptfield{Memory-use rules}Treat each retrieved card as one local page decision, never as a complete shopping plan. Match its task family, page phase, product attributes, and current available action. A matching success card may support the listed action. A failure card describes a pattern to avoid and a local correction; never copy its failed action. Ignore a card when these cues conflict.
\par\smallskip
\promptkey{MEMORY DECISION EVIDENCE:} \promptvar{retrieved\_cards} Use it only when its cue and exact available action match the current page.
\promptfield{Action rules}Choose exactly one action, \texttt{search[<keywords>]} or \texttt{click[<button-or-ASIN>]}. Search with precise product keywords; on results select the matching ASIN; on a product page choose required options before Buy Now. Do not click description, features, or reviews. Do not invent a product, option, or button.
\par\smallskip
\promptkey{RECENT ACTION HISTORY:} \promptvar{recent\_history}\par\vspace{1pt}
\promptkey{Latest observation:} \promptvar{current\_page}\par\vspace{1pt}
\promptkey{Admissible actions:} \promptvar{admissible\_actions}
\promptfield{Output}Return exactly one complete action inside \texttt{<action>...</action>}. Output no explanation.
\end{tcolorbox}

\begin{tcolorbox}[appendixpromptws,title={\webmark\hspace{.4em}(g) WebShop / 2. Memory extraction\hfill\textit{Role: User}}]
\promptfield{Role and task}You are \texttt{CriticalStepLocator}. Review the WebShop trajectory and return JSON only, using the top-level key \texttt{candidates}. Select at most 5 consequential decisions, ranked by \texttt{importance\_score}.
\promptfield{Output schema}Each item contains \texttt{step\_id}, \texttt{c\_t}, \texttt{critical\_role}, \texttt{importance\_score}, \texttt{importance\_type}, \texttt{why\_critical}, and \texttt{expected\_fix\_direction}. Set \texttt{critical\_role} to \texttt{success\_maintain}, \texttt{failure\_correction}, or \texttt{uncertain}.
\promptfield{Hindsight credit}Independently score each selected source decision with \texttt{c\_t} in $[-1, 1]$, using the goal, trajectory, and episode feedback. Assign a positive score to a local action that should be reinforced and a negative score to an action that should be corrected. Set \texttt{c\_t} to 0 and \texttt{critical\_role} to \texttt{uncertain} when the evidence does not support either role. For nonzero \texttt{c\_t}, its sign must agree with \texttt{critical\_role}: positive means \texttt{success\_maintain} and negative means \texttt{failure\_correction}. Judge the local step rather than copying the whole-episode outcome. Keep \texttt{importance\_score} as the separate candidate-ranking score.
\promptfield{Selection rules}Prefer the earliest correctable wrong decision in a failed episode, such as an irrelevant search, a mismatched ASIN, an omitted required option, or a premature Buy Now click. Retain successful choices whose product and option constraints match the instruction. Avoid routine page navigation, duplicate observations, future-only information, and whole-task solutions.
\par\smallskip
\promptkey{EPISODE OUTCOME:} \promptvar{episode\_outcome}\quad $\cdot$\quad \promptkey{GOAL:} \promptvar{goal}\par\vspace{1pt}
\promptkey{TRAJECTORY:} \promptvar{trajectory\_json}
\end{tcolorbox}

\begin{tcolorbox}[appendixpromptws,title={\webmark\hspace{.4em}(h) WebShop / 3. Hint construction\hfill\textit{Role: User}}]
\promptfield{Role and task}You are \texttt{ANCHOR write\_hint}. Return JSON only with \texttt{hint\_text}, \texttt{hint\_type}, and \texttt{leakage\_risk}.
\promptfield{Guidance rules}Write one short, local check-only hint for the selected WebShop decision. For a \texttt{failure\_correction}, identify the observed choice to avoid and the immediate action or attribute check that corrects it. For a \texttt{success\_maintain}, stabilize the observed matching action. Use only the current page, instruction, and admissible actions; do not reveal future pages, recommend an unavailable product, or give a complete shopping solution.
\par\smallskip
\promptkey{GOAL:} \promptvar{goal}\quad $\cdot$\quad \promptkey{CANDIDATE:} \promptvar{candidate\_json}\quad $\cdot$\quad \promptkey{STATE:} \promptvar{state\_json}
\end{tcolorbox}

\begin{tcolorbox}[appendixpromptws,title={\webmark\hspace{.4em}(i) WebShop / 4. Paired replay\hfill\textit{Role: User}}]
\promptkey{TASK:} \promptvar{goal}\quad $\cdot$\quad \promptkey{TASK FAMILY:} \promptvar{task\_family}\quad $\cdot$\quad \promptkey{PHASE GUIDANCE:} \promptvar{current\_phase}
\promptfield{Memory-use rules}Treat each retrieved card as one local page decision, never as a complete shopping plan. Match its task family, page phase, product attributes, and current available action. A matching success card may support the listed action. A failure card describes a pattern to avoid and a local correction; never copy its failed action. Ignore a card when these cues conflict.
\par\smallskip
\promptkey{MEMORY DECISION EVIDENCE:} \promptvar{retrieved\_cards} Use it only when its cue and exact available action match the current page.\par\vspace{1pt}
\promptkey{CANDIDATE GUIDANCE:} \promptvar{candidate\_hint} Apply this guidance only when it matches the current page and an available action.
\promptfield{Action rules}Choose exactly one action, \texttt{search[<keywords>]} or \texttt{click[<button-or-ASIN>]}. Search with precise product keywords; on results select the matching ASIN; on a product page choose required options before Buy Now. Do not click description, features, or reviews. Do not invent a product, option, or button.
\par\smallskip
\promptkey{RECENT ACTION HISTORY:} \promptvar{recent\_history}\par\vspace{1pt}
\promptkey{Latest observation:} \promptvar{current\_page}\par\vspace{1pt}
\promptkey{Admissible actions:} \promptvar{admissible\_actions}
\promptfield{Output}Return exactly one complete action inside \texttt{<action>...</action>}. Output no explanation.
\end{tcolorbox}

\begin{tcolorbox}[appendixpromptws,title={\webmark\hspace{.4em}(j) WebShop / 5. Card construction\hfill\textit{Role: User}}]
\promptfield{Role and task}You are \texttt{ANCHOR MemoryCardBuilder}. Return JSON only. Build a reusable step-level WebShop card from the verified decision, not a complete shopping plan.
\promptfield{Role mapping}Map \texttt{success\_maintain} to \texttt{success\_memory} and \texttt{failure\_correction} to \texttt{fail\_memory}; the step role is independent of the whole-episode outcome.
\promptfield{Card content}Store \texttt{VERIFIED HINT} verbatim as \texttt{memory\_guidance}. Add shopping phase, product attributes, option constraints, retrieval keys, source metadata, confidence, and support count in separate fields.
\promptfield{Guidance preservation}Do not paraphrase, summarize, or extend \texttt{memory\_guidance}. Generalize product-specific names or ASINs only in retrieval metadata, leaving the tested hint unchanged. Exclude future-only information from added metadata.
\par\smallskip
\promptkey{VERIFIED HINT:} \promptvar{candidate\_hint}\quad $\cdot$\quad \promptkey{FAMILY:} \promptvar{task\_family}\quad $\cdot$\quad \promptkey{GOAL:} \promptvar{goal}\par\vspace{1pt}
\promptkey{CANDIDATE:} \promptvar{candidate\_json}\quad $\cdot$\quad \promptkey{STATE:} \promptvar{state\_json}\quad $\cdot$\quad \promptkey{POLICY\_DIFF:} \promptvar{policy\_diff\_json}\quad $\cdot$\quad \promptkey{TASK\_PLAN:} \promptvar{task\_plan\_json}
\end{tcolorbox}

\section{Case Studies}
\label{app:cases}

The ALFWorld cases illustrate success reinforcement, failure suppression, and memory retrieval during execution. Paired probes compare action-verb probabilities at fixed source states.

\subsection{Similar Memories, Different Local Effects}
\label{app:case-navigation}

\begin{figure}[!htbp]
\centering
\includegraphics[width=.88\linewidth]{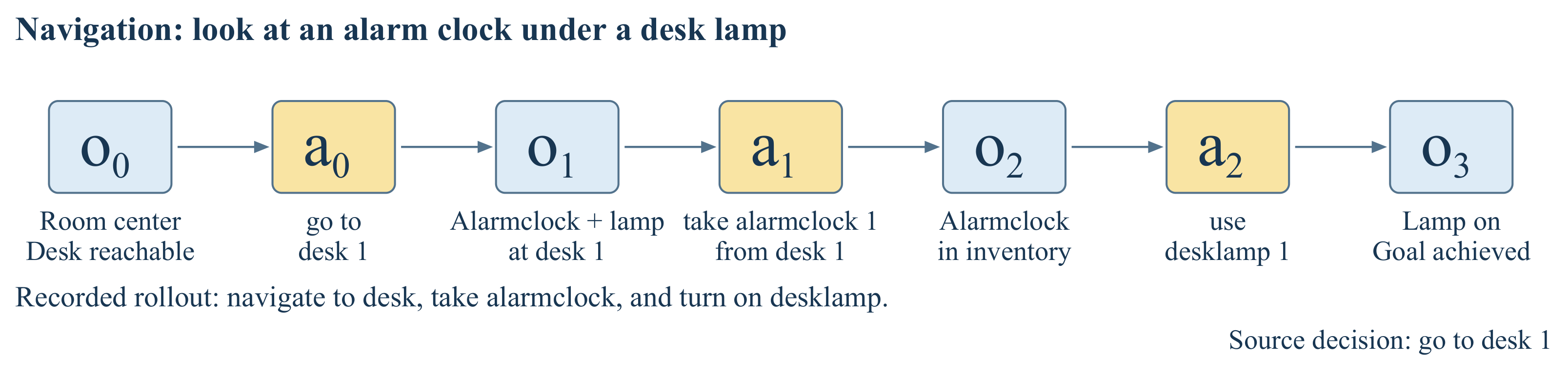}
\par\vspace{1.5pt}
\includegraphics[width=.60\linewidth]{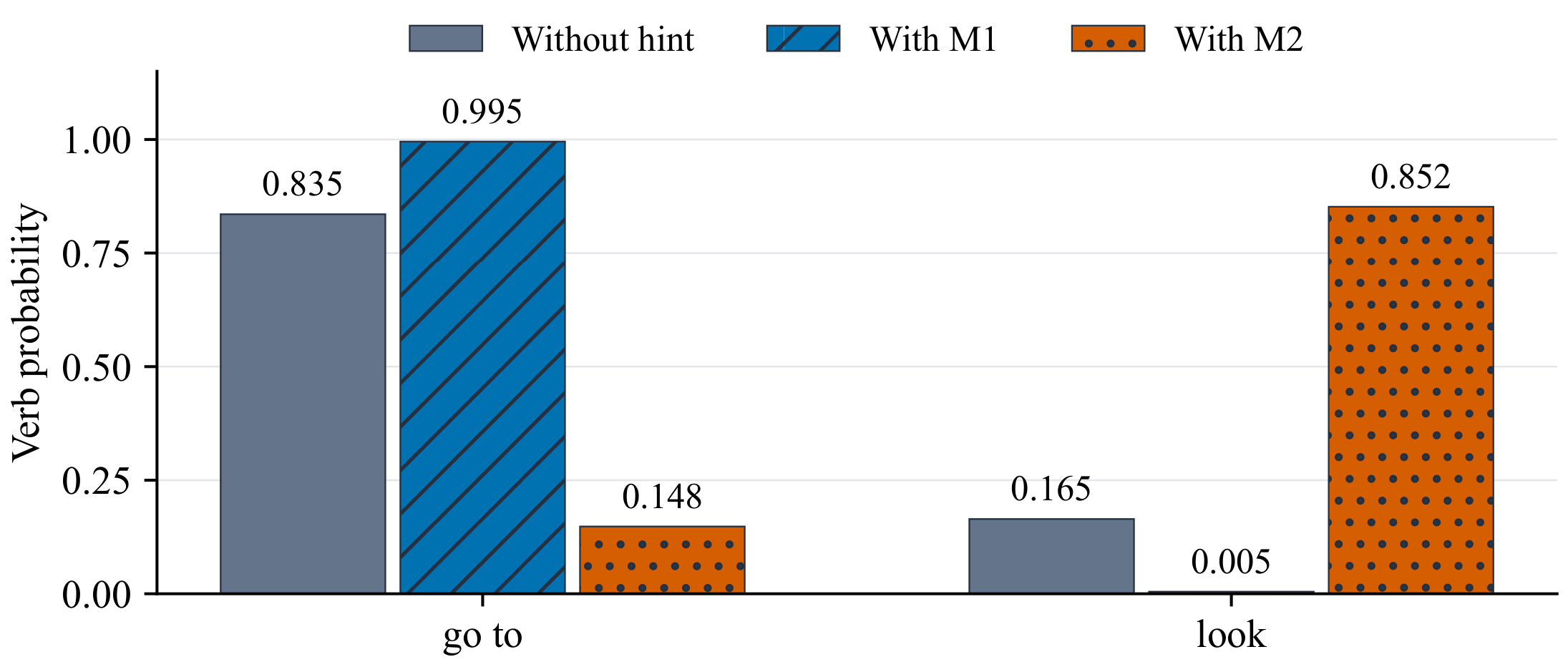}
\caption{\textbf{Navigation: the same source state, opposite memory effects.} The recorded rollout goes to desk 1, takes the alarm clock, and turns on the lamp to complete the task. Replay tests the initial navigation decision: M1 reinforces \texttt{go to}; M2 suppresses it and favors \texttt{look} (remaining verb mass $<4\times10^{-5}$).}
\label{fig:case-navigation}
\label{fig:case-navigation-probe}
\end{figure}

\paragraph{Source trajectory.}
The task is to look at an alarm clock under a desk lamp. At the initial decision, the agent is in the middle of the room and selects \texttt{go to desk 1}. The next observation reveals the alarm clock and lamp on that desk (Figure~\ref{fig:case-navigation}). The recorded source verb \texttt{go to} is also the baseline probe's top choice, with probability 0.835.

\paragraph{Local comparison.}
The two records share the initial observation, admissible actions, source action, and baseline distribution. Their hints differ in whether navigation is the action to take or an action already completed (Figure~\ref{fig:case-navigation-probe}).

\par\smallskip\noindent
\begin{minipage}[t]{.485\textwidth}
\vspace{0pt}
\begin{tcolorbox}[appendixsuccess,equal height group=nav-hints,
  top=3pt,bottom=3pt,before skip=2pt,after skip=2pt,
  title={\successmark\hspace{.4em}M1: Support navigation (\textbf{Pass})}]
The action `\texttt{go to desk 1}' successfully navigates you to the location containing the target objects. Proceed to interact with the alarmclock as specified in the goal.
\end{tcolorbox}
\end{minipage}\hfill
\begin{minipage}[t]{.485\textwidth}
\vspace{0pt}
\begin{tcolorbox}[appendixfailure,equal height group=nav-hints,
  top=3pt,bottom=3pt,before skip=2pt,after skip=2pt,
  title={\failuremark\hspace{.4em}M2: Premature completion (\textbf{Reject})}]
You have successfully navigated to desk 1. Verify that the alarmclock is present in your current view before proceeding.
\end{tcolorbox}
\end{minipage}
\par\smallskip

\paragraph{Interpretation.}
M1 reinforces the successful source decision, increasing its verb probability by 0.160. M2 instead describes navigation as completed at a pre-navigation state: it reduces \texttt{go to} by 0.687 and raises \texttt{look} to 0.852. Both shifts exceed the success-role magnitude threshold (Table~\ref{tab:case-navigation-local}), but only M1 reinforces the source action. The larger shift from M2 therefore fails the directional test.

\subsection{A Local Correction for a Failed Heating Trajectory}
\label{app:case-failure}

\begin{figure}[!htbp]
\centering
\includegraphics[width=.84\linewidth]{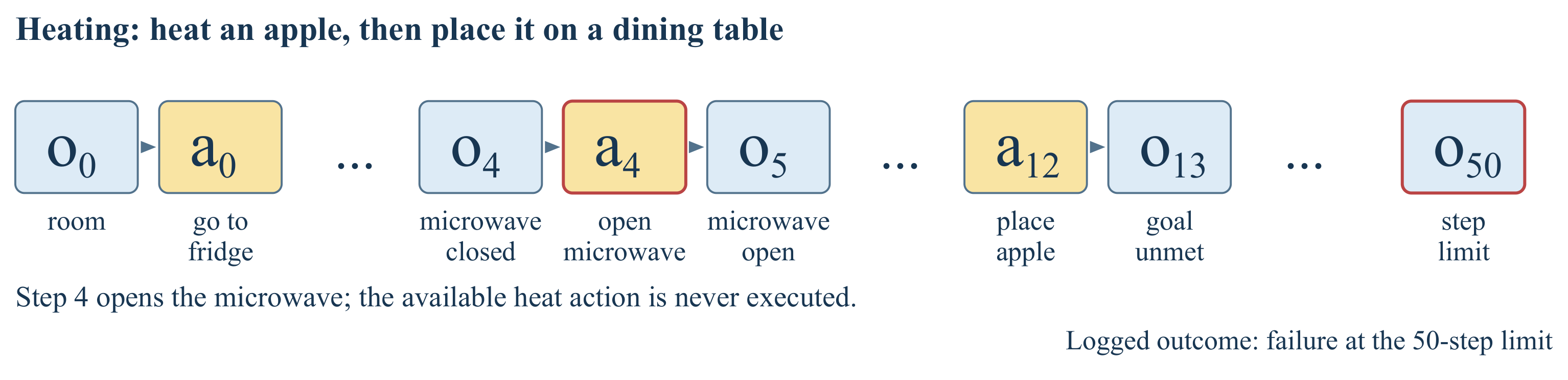}
\par\vspace{1pt}
\includegraphics[width=.54\linewidth]{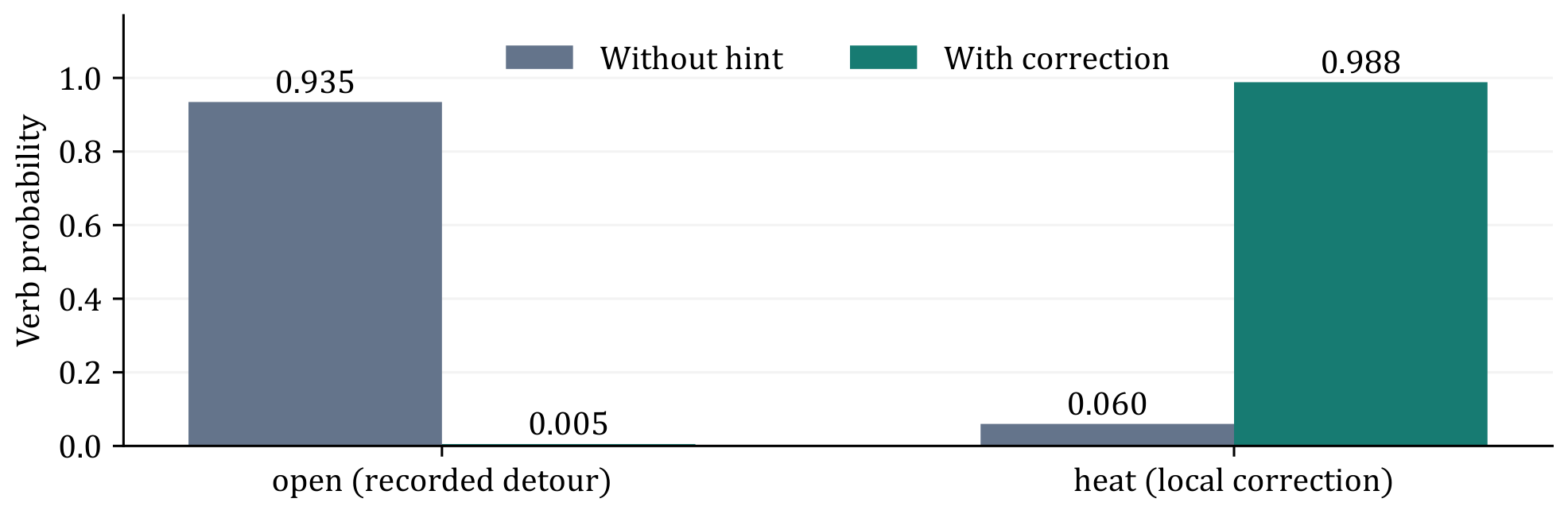}
\vspace{-3pt}
\caption{\textbf{Heating: correcting a failed source decision.} The agent opens the microwave at step 4 and never executes \texttt{heat}, exhausting the 50-step limit (dots omit intervening steps; zero-based indices). The extracted hint suppresses \texttt{open} and reinforces \texttt{heat}, reversing their ranking.}
\label{fig:case-failure}
\label{fig:case-failure-probe}
\end{figure}

\paragraph{Source trajectory.}
The goal is to heat an apple and place it on a dining table. After collecting the apple and reaching the microwave, the recorded agent opens the microwave, moves the apple inside, closes it, and later retrieves and places it. It never executes the environment's explicit \texttt{heat} action. Subsequent repeated actions and empty outputs exhaust the 50-step limit. Figure~\ref{fig:case-failure} shows selected observations and actions in their recorded order.

\Needspace{10\baselineskip}
\paragraph{Localizer record.}
Localizer identifies the step-4 detour in \texttt{train-00058-000}. The panels below pair the localized critical-step metadata with the extracted correction hint.

\par\smallskip\noindent
\begin{minipage}[t]{.54\textwidth}
\vspace{0pt}
\begin{tcolorbox}[appendixrecord,equal height group=heat-case-row,
  top=3pt,bottom=3pt,before skip=2pt,after skip=2pt,
  title={\localizermark\hspace{.4em}Localized critical step (\texttt{train-00058-000} / step 4)}]
\raggedright
\textbf{Source action:} \texttt{open microwave 1}\par
\textbf{Admissible:} \texttt{open microwave 1}; \texttt{heat apple 1 with microwave 1}\par
\textbf{Role / score:} \texttt{failure\_correction} (score $1.1$, \texttt{heat protocol detour})\par\smallskip
\textbf{Why critical:} Exact target \texttt{heat} action was available, but the rollout detoured.\par
\textbf{Expected fix:} Execute the exact target \texttt{heat} action immediately when listed.
\end{tcolorbox}
\end{minipage}\hfill
\begin{minipage}[t]{.435\textwidth}
\vspace{0pt}
\begin{tcolorbox}[appendixfailure,fontupper=\appendixfigfont\raggedright,
  fonttitle=\bfseries\appendixfigfont,equal height group=heat-case-row,
  top=3pt,bottom=3pt,before skip=2pt,after skip=2pt,
  title={\failuremark\hspace{.4em}Extracted failure-correction hint}]
Do not open the microwave first; the admissible actions list includes a direct command to heat the apple with the microwave, which should be executed immediately to achieve the goal.
\end{tcolorbox}
\end{minipage}
\par\smallskip

\paragraph{Local comparison.}
At step 4, the hint reduces the recorded source verb \texttt{open} from 0.935 to 0.005 and raises \texttt{heat} from 0.060 to 0.988 (Figure~\ref{fig:case-failure-probe}). JS divergence is 0.806 bits, and the recorded checker accepts the correction. This example demonstrates the intended failure-suppression direction at the replay state.

\Needspace{8\baselineskip}
\subsection{Memory Cards}
\label{app:case-card}
\label{app:case-fail-card}
Figure~\ref{fig:case-cards} groups the stored examples by role. Success-maintain memories support target acquisition, state modification, and tool activation; failure corrections address unnecessary appliance interactions, unrelated objects, and spatial-goal errors.

\begin{figure}[t!]
\centering
\makebox[.485\textwidth][l]{\normalsize\successmark\hspace{.45em}\textbf{\textcolor{rq3!70!black}{Success-maintain memories}}}\hfill
\makebox[.485\textwidth][l]{\normalsize\failuremark\hspace{.45em}\textbf{\textcolor{rq4!75!black}{Failure corrections}}}\par\vspace{2pt}
\begin{minipage}[t]{.485\textwidth}
\vspace{0pt}
\begin{tcolorbox}[appendixsuccess,fontupper=\appendixfigfont\raggedright,fonttitle=\bfseries\appendixfigfont,
  top=2.5pt,bottom=2.5pt,before skip=0pt,after skip=0pt,
  title={S1: Acquire the target object},equal height group=case-card-row-1]
\textbf{Source hint:} The alarmclock is currently on desk 1. To proceed with the task, you need to add it to your inventory.
\par\vspace{2pt}{\color{rq3!35}\hrule}\vspace{2pt}
{\footnotesize\textbf{Source action:} \texttt{take alarmclock 1 from desk 1}\quad $\cdot$\quad \textbf{Family:} \texttt{look\_at\_obj}}
\end{tcolorbox}
\end{minipage}\hfill
\begin{minipage}[t]{.485\textwidth}
\vspace{0pt}
\begin{tcolorbox}[appendixfailure,fontupper=\appendixfigfont\raggedright,fonttitle=\bfseries\appendixfigfont,
  top=2.5pt,bottom=2.5pt,before skip=0pt,after skip=0pt,
  title={F1: Avoid opening before direct heating},equal height group=case-card-row-1]
\texttt{memory\_guidance:} Do not open the microwave first; the admissible actions list includes a direct command to heat the apple with the microwave, which should be executed immediately to achieve the goal.
\par\vspace{2pt}{\color{rq4!35}\hrule}\vspace{2pt}
{\footnotesize\textbf{Action:} \texttt{open microwave 1}\ $\cdot$\ \textbf{ID:} \texttt{self-local-00058-004}}
\end{tcolorbox}
\end{minipage}\par\vspace{2.5pt}
\begin{minipage}[t]{.485\textwidth}
\vspace{0pt}
\begin{tcolorbox}[appendixsuccess,fontupper=\appendixfigfont\raggedright,fonttitle=\bfseries\appendixfigfont,
  top=2.5pt,bottom=2.5pt,before skip=0pt,after skip=0pt,
  title={S2: Complete cleaning before placement},equal height group=case-card-row-2]
When a task requires an object to be in a specific state (e.g., `clean') before storage, execute the state-modifying action (e.g., `clean') at the appropriate facility (e.g., `sink') immediately upon arrival, rather than proceeding to the storage location or performing unrelated navigation.
\par\vspace{2pt}{\color{rq3!35}\hrule}\vspace{2pt}
{\footnotesize\textbf{Action:} \texttt{clean apple 3 with sinkbasin 1}\ $\cdot$\ \textbf{ID:} \texttt{self-local-00062-003}}
\end{tcolorbox}
\end{minipage}\hfill
\begin{minipage}[t]{.485\textwidth}
\vspace{0pt}
\begin{tcolorbox}[appendixfailure,fontupper=\appendixfigfont\raggedright,fonttitle=\bfseries\appendixfigfont,
  top=2.5pt,bottom=2.5pt,before skip=0pt,after skip=0pt,
  title={F2: Ignore an unrelated object},equal height group=case-card-row-2]
When an opened container reveals an object that does not match the target item required for the current heating or placement goal, do not interact with the found object. Instead, close the container and resume the search for the correct target item to avoid unnecessary detours.
\par\vspace{2pt}{\color{rq4!35}\hrule}\vspace{2pt}
{\footnotesize\textbf{Action:} \texttt{take pot 1 from cabinet 2}\ $\cdot$\ \textbf{ID:} \texttt{self-local-00022-006}}
\end{tcolorbox}
\end{minipage}\par\vspace{2.5pt}
\begin{minipage}[t]{.485\textwidth}
\vspace{0pt}
\begin{tcolorbox}[appendixsuccess,fontupper=\appendixfigfont\raggedright,fonttitle=\bfseries\appendixfigfont,
  top=2.5pt,bottom=2.5pt,before skip=0pt,after skip=0pt,
  title={S3: Activate the tool before examination},equal height group=case-card-row-3]
When a task requires examining an object with a specific tool, and the tool is present but inactive, the critical step is to activate the tool using the `use' action before attempting the examination. This activation is a prerequisite for the `examine with' condition to be satisfied.
\par\vspace{2pt}{\color{rq3!35}\hrule}\vspace{2pt}
{\footnotesize\textbf{Action:} \texttt{use desklamp 1}\ $\cdot$\ \textbf{ID:} \texttt{self-local-00048-003}}
\end{tcolorbox}
\end{minipage}\hfill
\begin{minipage}[t]{.485\textwidth}
\vspace{0pt}
\begin{tcolorbox}[appendixfailure,fontupper=\appendixfigfont\raggedright,fonttitle=\bfseries\appendixfigfont,
  top=2.5pt,bottom=2.5pt,before skip=0pt,after skip=0pt,
  title={F3: Respect the spatial goal},equal height group=case-card-row-3]
When a goal specifies an object located relative to another object (e.g., `under X'), do not interact with instances of that object found elsewhere. Instead, navigate to the reference object and verify the spatial relationship before acting. Avoid taking the object if the goal is merely to observe it in situ.
\par\vspace{2pt}{\color{rq4!35}\hrule}\vspace{2pt}
{\footnotesize\textbf{Action:} \texttt{take alarmclock 2 from desk 1}\ $\cdot$\ \textbf{ID:} \texttt{self-local-00096-001}}
\end{tcolorbox}
\end{minipage}
\caption{\textbf{Memory cards grouped by source role.} Left: success-maintain memories S1--S3. Right: failure corrections F1--F3. Guidance text and source metadata are shown together; F1 preserves the tested heating hint verbatim.}
\label{fig:case-cards}
\end{figure}
\par\smallskip

\subsection{Inference-Time Memory Injection}
\label{app:inference-injection}

\paragraph{Paired source-state probes.}
Table~\ref{tab:case-injection-local} compares source-verb probabilities at fixed saved contexts. S1 increases \texttt{take} by 0.265 and retains it as the top choice; F1 decreases \texttt{open} by 0.929 and makes \texttt{heat} the top choice. Both satisfy the admission rule.

\par\smallskip\noindent
\begin{minipage}{\textwidth}
\centering
\appendixfigfont
\anchortablestyle
\renewcommand{\arraystretch}{1.08}
\captionof{table}{\textbf{Local effects of memory injection.} Paired source-state probes across navigation hints, success reinforcement, and failure suppression ($p_0\rightarrow p_1$: source-verb probability before $\rightarrow$ after injection; source verb is baseline top choice).}
\label{tab:case-navigation-local}
\label{tab:case-injection-local}
\begin{tabular*}{\textwidth}{@{\extracolsep{\fill}}llrrll@{}}
\toprule
\tableheader{Hint / source record} & \tableheader{Source verb} & \tableheader{$p_0 \rightarrow p_1$} & \tableheader{JS (bits)} & \tableheader{Injected top verb} & \tableheader{Gate check}\\
\midrule
\rowcolor{groupgray}
\multicolumn{6}{@{}l@{}}{\tablegroup{compass}{Navigation hints at the same source state}}\\
M1 / \texttt{eval-00006-000} & \texttt{go to} & $.835 \rightarrow .995$ & $.074$ & \texttt{go to} & \appsuccessbadge\\
M2 / \texttt{eval-00007-000} & \texttt{go to} & $.835 \rightarrow .148$ & $.375$ & \texttt{look} & \apprejectbadge\\
\addlinespace[1.5pt]
\rowcolor{groupgray}
\multicolumn{6}{@{}l@{}}{\tablegroup{check-circle}{Success reinforcement and failure suppression}}\\
S1 / \texttt{train-00012-001} & \texttt{take} & $.735 \rightarrow 1.000$ & $.146$ & \texttt{take} & \appsuccessbadge\\
F1 / \texttt{train-00058-000} & \texttt{open} & $.935 \rightarrow .005$ & $.806$ & \texttt{heat} & \appsuccessbadge\\
\bottomrule
\end{tabular*}
\end{minipage}
\par\smallskip

\paragraph{Source records.}
\label{app:case-provenance}
Table~\ref{tab:case-injection-local} lists the source records. The navigation probes share the initial state; the heating record links to F1 (\texttt{self-local-00058-004}), which preserves the tested hint. S1 quotes its source hint; S2, S3, F2, and F3 reproduce stored guidance. All probes use the complete saved context, $L=1$, and base-2 JS (Appendix~\ref{app:prompts}). Trajectory excerpts retain the recorded action order. Table~\ref{tab:case-later-retrieval} identifies the later retrieval episodes and steps.

\par\smallskip\noindent
\begin{minipage}{\textwidth}
\centering
\appendixfigfont
\anchortablestyle
\setlength{\tabcolsep}{4pt}
\renewcommand{\arraystretch}{1.08}
\captionof{table}{\textbf{Later retrievals in the construction run.} Stored-card retrievals and executed actions during memory construction.}
\label{tab:case-later-retrieval}
\begin{tabular*}{\textwidth}{@{\extracolsep{\fill}}lllr@{}}
\toprule
\tableheader{Episode / step} & \tableheader{Retrieved card} & \tableheader{Executed action} & \tableheader{Feedback (abridged)}\\
\midrule
\rowcolor{groupgray}
\multicolumn{4}{@{}l@{}}{\tablegroup{database}{Constructed-bank retrievals across online episodes}}\\
104 / 4 & \texttt{self-local-00062-003} & \texttt{clean apple 3 with sinkbasin 1} & Apple 3 is cleaned.\\
128 / 6 & \texttt{self-local-00062-003} & \texttt{clean apple 2 with sinkbasin 1} & Apple 2 is cleaned.\\
68 / 9 & \texttt{self-local-00062-006} & \texttt{move apple 1 to fridge 1} & Apple 1 is placed in the fridge.\\
\bottomrule
\end{tabular*}
\end{minipage}
\par\smallskip

\paragraph{A later retrieval during execution.}
At step 4 of training episode 104, the agent is at \texttt{sinkbasin 1} with apple 3 in inventory. Its single retrieved card, \texttt{self-local-00062-003}, supplies this local instruction:

\begin{tcolorbox}[appendixsuccess,top=3pt,bottom=3pt,before skip=3pt,after skip=3pt,
  title={\bankmastermark\hspace{.45em}Retrieved cue and executed action (Episode 104, Step 4)}]
\textbf{Injected cue:} \texttt{NEXT ACTION IF MATCHED: clean apple 3 with sinkbasin 1}\par\vspace{1.5pt}
\textbf{Available \& executed action:} \texttt{clean apple 3 with sinkbasin 1}\par\vspace{1.5pt}
\textbf{Next observation:} ``You clean the apple 3 using the sinkbasin 1.''\quad $\cdot$\quad \textbf{Episode outcome:} \textcolor{rq3!75!black}{\textbf{Success}}
\end{tcolorbox}

The same card guides another cleaning decision with a different apple; a second card supports final placement. In all three steps in Table~\ref{tab:case-later-retrieval}, \texttt{NEXT ACTION IF MATCHED} is admissible and equals the executed action. All three episodes are logged as successful: the paired probes isolate a local policy change, and the later traces record retrieval followed by the matching action and observation.

\end{document}